\documentclass[runningheads,a4paper]{llncs}
\usepackage{amssymb}
\usepackage{graphicx}
\usepackage{epsfig}
\usepackage{amsmath}
\usepackage{amssymb}
\usepackage{subfigure}
\usepackage{float}
\usepackage{comment}
\usepackage{times}
\usepackage{url}
\urldef{\mailsa}\path|{fariborztaherkhani@gmail.com|
\urldef{\mailsb}\path|Jeremy.Dawson@mail|
\urldef{\mailsc}\path|nasser.nasrabadi@mail}@wvu.edu|

\begin{document}

\mainmatter  

\title{Deep Sparse Band Selection for Hyperspectral Face Recognition}

\titlerunning{Lecture Notes in Computer Science: Authors' Instructions}

%
%
\author{Fariborz Taherkhani%
\and Jeremy Dawson\and Nasser M. Nasrabadi}
\authorrunning{Lecture Notes in Computer Science: Authors' Instructions}

\institute{West Virginia University\\
\{ft0009@mix, jeremy.dawson@mail, nasser.nasrabadi@mail\}.wvu.edu}

%
%

\toctitle{Lecture Notes in Computer Science}
\tocauthor{Authors' Instructions}
\maketitle
\begin{abstract}
Hyperspectral imaging systems collect and process information from specific wavelengths across the electromagnetic spectrum.  The fusion of multi-spectral bands in the visible spectrum has been exploited to improve face recognition performance over all the conventional broad band face images. In this book chapter, we propose a new Convolutional Neural Network (CNN) framework which adopts a structural sparsity learning technique to select the optimal spectral bands to obtain the best face recognition performance over all of the spectral bands. Specifically, in this method,  images from all bands are fed to a CNN, and the convolutional filters in the first layer of  the CNN are then regularized by employing a group Lasso algorithm to zero out the redundant bands during the training of the network. Contrary to other methods which usually select the useful bands manually or in a greedy fashion, our method selects the optimal  spectral bands automatically to achieve the best face recognition performance over all spectral bands. Moreover, experimental results demonstrate that our method outperforms state of  the art band selection methods for face recognition on several publicly-available hyperspectral face image datasets. 
\end{abstract}

\section{Introduction}
In recent year, hyperspectral imaging has attracted much attention due to the decreasing  cost of hyperspectral cameras used for image accusation  \cite{allen2016overview}.
A hyperspectral image consists of many narrow spectral bands within the visible spectrum and beyond. This data  is
structured as a hyperspectral "cube", with x and y coordinates making
up the imaging pixels and the z coordinate the imaging wavelength,
which, in the case of facial imaging, results in 
several co-registered face images captured  at
varying wavelengths.  Hyperspectral imaging has provided new opportunities for improving the performance of  different imaging tasks, such as face recognition in biometrics, that exploits the spectral characteristics of facial tissues to increase the inter-subject differences \cite{pan2003face}. It has been demonstrated that, by adding the extra spectral dimension, the size of the feature space representing a face image is increased  which results in  a larger inter-class features differences between subjects for face recognition. Beyond the surface appearance, spectral measurements in the infra-red (i.e., 700 $n m$  to 1000 $nm $) can penetrate  the subsurface tissue which can notably produce different biometric features for each subject \cite{uzair2015hyperspectral}.

 A hyperspectral imaging camera simultaneously measures hundreds of adjacent spectral bands with a small spectral resolution (e.g.,  10 $n m$). For example, AVIRIS hyperspectral imaging includes $224$ spectral bands from  400 $nm$ to 2500 $nm$ \cite{kruse2002comparison}. Such a large number of bands implies high-dimensional data which remarkably influences the performance of face recognition. This is because, a redundancy exists between spectral bands, and  some bands may  hold less discriminative information than others. Therefore, it is advantageous to discard bands which carry little or no discriminative information during the recognition task. To deal with this problem, many band selection approaches have been proposed in order to choose the optimal and  informative bands for face recognition. Most of these methods, such as those presented in  \cite{pan2009comparison}, are based on dimensionality reduction, but in an ad-hoc fashion.  These methods, however, suffer from a lack of comprehensive and consolidated evaluation due to a) the small number of subjects used during the testing of the methods, and b)  lack of publicly available datasets  for comparison. Moreover, these studies  do not compare the performance of their algorithms comprehensively with other face recognition approaches that can be used for this challenge with some modifications \cite{uzair2015hyperspectral}. 
 
The development of  hyperspectral cameras has introduced many useful techniques that merge spectral and spatial information. Since hyperspectral cameras have become more readily available, computational
approaches introduced initially for remote sensing
challenges have been leveraged to other application such as biomedical applications. Considering the vast person-to-person spectral variability for different types of tissue, hyperspectral imaging has
the power to enhance the capability of automated systems
for human re-identification.
Recent face recognition protocols essentially apply spatial
discriminants that are based on geometric facial features \cite{kruse2002comparison}. Many of these protocols have provided promising results
on databases captured under controlled conditions.
However, these methods often indicate significant performance
drop in the presence of variation in face
orientation \cite{pan2003face,ryer2012quest}. 

The work in \cite{gross2001quo}, for instance, indicated that
there is significant drop in the performance of recognition 
for images of faces which are rotated more than 32 degrees
from a frontal image that is used to train the model. Furthermore, in  \cite{gross2004appearance}, which uses a light-fields model for
pose-invariant face recognition, provided well recognition
results for probe faces which are rotated more than 60 degrees
from a gallery face. The method, however, requires the
manual determination of the 3D transformation  to
register face images. The methods that use geometric features
can also perform poorly if subjects are imaged sacross varying spans of time. For instance, recognition performance can
decrease by a maximum of 20 \% if imaging sessions
are separated by a two week interval \cite{gross2001quo}. Partial face occlusion also usually results in poor performance. An approach \cite{martinez2002recognizing} that divided the face images into regions for isolated analysis
can tolerate up to $1/6$ face occlusion without a decrease in matching
accuracy. Thermal infrared imaging provides an alternative
imaging modality that has been leveraged for face recognition \cite{wilder1996comparison}. However, algorithms based on thermal images utilize spatial features and have difficulty recognizing faces when presented with images containgin pose variation.

A 3D morphable face approach has been introduced for face recognition across variant poses \cite{blanz2002face}.
This method has provided a good performance on a 68-subject dataset. However, this method is
 currently computationally intensive and requires significant manual
intervention. Many of the limitations of current face recognition methods can be overcome by leveraging spectral information. The interaction of light with human tissue has been explored
comprehensively by many works \cite{anderson1981optics} which consider the spectral properties of tissue. The epidermal and
dermal layers of human skin are essentially a scattering medium
that consists of several pigments such as hemoglobin, melanin, bilirubin, and carotene. Small changes in the distribution of
these pigments cause considerable changes in the skin’s
spectral reflectance \cite{edwards1939pigments} . For
instance, the impacts are large enough to enable algorithms for the automated separation
of melanin and hemoglobin from RGB images \cite{tsumura1999independent}. Recent
work \cite{angelopoulo2001multispectral} has calculated skin reflectance spectra over the
visible wavelengths and introduced algorithms for the spectra.

\section{Related Work}

 \subsection{Hyperspectral Imaging Techniques}
There are three common techniques used to construct a hyperspectral image: spatial scanning, spectral scanning, or snapshot imaging. These techniques will be described in detail in the following sections.

Spatial scan systems capture each spectral band along a single dimension as a scanned composite image of the object or area being viewed.  The scanning aspect of these systems describes the narrow imaging field of view  (e.g., a 1xN pixel
array)  of the system.  The system creates images using an optical slit to allow only a thin strip of the image to pass through a prism or grating that then projects the diffracted scene onto an imaging sensor.  By limiting the amount of scene (i.e. spatial) information into the system at any given instance, most of the imaging sensor area can be utilized to capture spectral information. This reduction in spatial resolution allows for simultaneous capture of data at a higher spectral resolution.  This data capture technique is a practical solution for applications where a scanning operation is possible, specifically for airborne mounted systems that image the ground surface as an aircraft flies overhead.  Food quality inspection is another successful application of these systems, as they can rapidly detect defective or unhealthy produce on a production or sorting line.  While this technique provides both high spatial and spectral resolution, line scan Hyperspectral Imaging  Systems (HSIs) are highly susceptible to changes the morphology of the target.  This means the system must be fixed to a steady structure as the subject passes through its linear field of view or, that the subject remains stationary as the imaging scan is conducted.

\begin{figure*}
\centering
\subfigure[]{\includegraphics[trim={0 0 0 0}, scale=0.29]{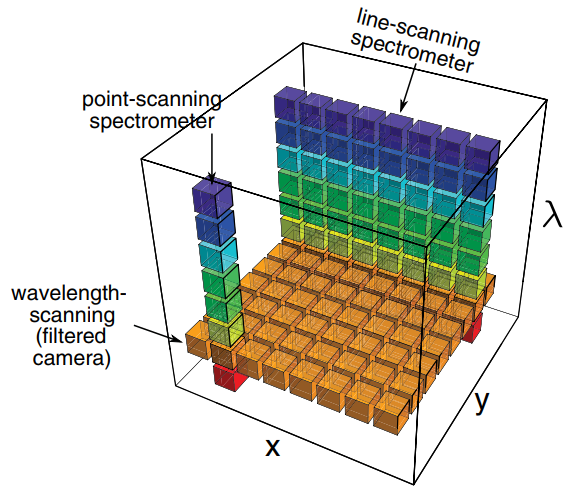}}
\subfigure[]{\includegraphics[scale=0.29]{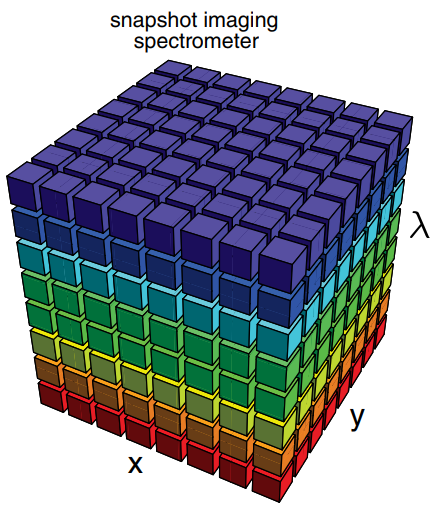}}
\caption{Building the spectral data cube in both line scan and snapshot systems}
\label{Rank_LFW1}
\end{figure*}

HSIs, such as those employing an Acousto-Optical Tunable Filter (AOTF) or a Liquid Crystal Tunable Filter (LCTF), use tunable optical devices that allow specific wavelengths of electromagnetic radiation to pass through to a broadband camera sensor.  While the fundamental technology behind these tunable filters is different, their application achieves the same goal in a similar fashion by iteratively selecting the spectral bands of a subject that fall on the imaging sensor. Depending on the type of filter used, the integration time between the capture of each band can vary based upon the driving frequency of the tunable optics and the integration time of the imaging plane.  One limitation of scanning HSIs is that all bands in the data cube cannot be captured simultaneously. Fig. \ref{Rank_LFW1} (a) \cite{hagen2013review} illustrates a diagram which depicts the creation of the hyperspectral data cube by spatial and spectral scanning. 

In contrast to scanning methods, a snapshot hyperspectral camera can capture hyperspectral image data in which all wavelengths are captured instantly to create the hypercube, as shown in Fig. \ref{Rank_LFW1} (b) \cite{hagen2013review}. Snapshot hyperspectral technology is designed and built in configurations different from line scan imaging systems, often employing a prism to break up the light and causing the diffracted, spatially separated wavelengths to fall on different portions of the imaging sensor dedicated to collecting light energy from a specific wavelength. Software is used to sort the varying wavelengths of light falling onto different pixels into wavelength-specific groups. While conventional line-scan hyperspectral cameras build the data cube by scanning through various filtered wavelengths or spatial dimensions, the snapshot HSI acquires an image and the spectral signature at each pixel simultaneously. Snapshot systems have an advantage of faster measurement and higher sensitivity. However, one drawback is that the resolution is limited by down-sampling the light falling onto the imaging array into a smaller number of spectral channels.

\subsection{Spectral Face Recognition}
Most hyperspectral face recognition approaches are an extension of  typical face recognition methods which have been adjusted to this challenge. For example, each band of a hyperspectral image  can be considered  as a separate image, and as a result, gray-scale face recognition approaches  can be  applied to them. 

Considering a hyperspectral cube as a set of images, image-set classification approaches can be leveraged for this problem without using a dimensionality reduction algorithm \cite{uzair2015hyperspectral}. For example,  Pan \textit{et al.} \cite{pan2003face} used  31 spectral band signatures at manually-chosen landmarks  on face images which were captured  within the near infra-red spectrum.
 Their method provided high recognition accuracy under pose variations on a dataset which contains 1400 hyperspectral images from 200 people.  However, the method does not achieve the same promising results on the public hyperspectral datasets used in \cite{ryer2012quest}.
 
 Later on, Pan \textit{et al.} \cite{pan2009comparison}  incorporated  spatial and spectral information to improve the recognition results on  the same dataset. Robila \cite{robila2008toward} distinguished spectral signatures of different face locations by leveraging  spectral angle measurements. Their experiments are restricted to  a very small dataset which consists of only 8 subjects. Di \textit{et al.} \cite{di2010studies} projected the cube of hyperspectral images to a lower dimensional space by using a two-dimensional PCA method, and  then Euclidean  distance was calculated for face recognition.  Shen \textit{et al.} \cite{shen2012hyperspectral} used  Gabor-wavelets on hyperspectral data cubes to  generate 52 new cubes from each given cube. Then, they used an ad-hoc sub-sampling algorithm to reduce the large amount of data for face recognition.

 A wide variety of approaches  have been used to address the challenge of band selection for hyperspectral face recognition. Some of these methods are information-based methods \cite{bajcsy2004methodology}, transform-based methods \cite{chang1999joint}, search based methods \cite{melgani2004classification}, and other techniques which include maximization of a spectral angle mapper \cite{keshava2001best}, high-order moments \cite{du2003band}, wavelet analysis \cite{kaewpijit2003automatic}, and a scheme trading spectral for spatial resolution \cite{price1997spectral}. Nevertheless, there are still some challenges with  these approaches  due to presence of local-minima problems, difficulties for real-time implementation and high computational cost.
Hyperspectral imaging techniques for face recognition have provided promising results in the field of biometrics, overcoming challenges such as  pose variations, lighting variations, presentation attacks  and facial expression variations \cite{steiner2016reliable}. The fusion of narrow-band spectral images in the visible spectrum has been explored to enhance face recognition performance \cite{bouchech2014dynamic}. 
For example, Chang \textit{et al.} \cite{chang1999joint} have demonstrated that the fusion  of 25 spectral bands can surpass the performance of conventional broad band images for face recognition, mainly in cases where the training and testing images are collected under different types of illumination.

Despite  the new opportunities provided by hyperspectral imaging, challenges still exist due to low signal to noise ratios, high dimensionality,  and difficulty in data acquisition \cite{taherkhani2017restoring}. For example, hyperspectral images are usually stacked sequentially; hence,  subject movements, specifically blinking of the eyes, can lead to  band  misalignment. This misalignment causes  intra-class variations which cannot be compensated for by  adding  spectral dimension. Moreover, adding a spectral dimension makes the recognition task challenging due to the difficulty of choosing the required discriminative information. Furthermore,  the spectral dimension causes a curse of dimensionality concern, because the ratio  between the dimension of the data  and the number of  training  data becomes very large \cite{uzair2015hyperspectral}. 

Sparse dictionary learning has only  been extended to the hyperspectral image classification \cite{chen2013hyperspectral}. Sparse-based hyperspectral image classification methods usually rank the contribution of each band in the classification task, such that each band is approximated by a linear combination of a dictionary, which contains other band images. The sparse coefficients represent the contribution of each dictionary atom to the target band image, where the large coefficient shows that the band has significant contribution for classification, while the small coefficient indicates that the band has negligible contribution for classification.

 In recent years, deep learning methods have shown impressive learning ability in image retrieval \cite{talreja2019attribute,taherkhani2018facial,talreja2018using,talreja2018using}, generating images \cite{taherkhani2019matrix,kazemi2018unsupervised,kazemi2018unsupervised0}, security purposes \cite{talreja2017multibiometric,talreja2018biometrics} , image classification \cite{he_resnet_2016,krizhevsky_imagenet_2012,simonyan_very_deep_2014}, object detection \cite{Erhan_2014_CVPR,ren_faster_rcnn_2015}, face recognition \cite{soleymani2019defending,soleymani2018prosodic,soleymani2019adversarial, taherkhani2018deep,schroff2015facenet,taigman2015web,sun2014deep} and many other computer vision and biometrics tasks. 
  In addition to improving performance in computer vision and biometrics tasks, deep learning in combination with reinforcement learning methods was able to defeat the human champion in challenging games such as Go \cite{silver_2016_mastering}.
 CNN-based models have also been applied to hyperspectral image classification \cite{zhong2017deep}, band selection \cite{zhan2017hyperspectral,zhan2017new}, and hyperspectral face recognition \cite{sharma2016hyperspectral}. However, few of  these methods have provided  promising results for hyperspectral image classification due to a sub-optimal learning process caused by an insufficient amount of training data, and the use of comparatively small scale CNNs \cite{li2018novel}. 
\subsection{Spectral Band Selection}
Previous research on  band selection for face recognition usually works in an ad hoc fashion where  the combination of  different bands is evaluated to determine the best recognition performance. For instance, Di \textit{et al.} \cite{di2010studies} manually choose two disjoint  subsets of bands which are centered at 540 $nm$ and 580 $nm$ to examine their discrimination power. However, selecting the optimal bands manually may not be appropriate  because of the huge search space of many spectral bands.

In another case, Guo \textit{et al.} \cite{guo2012feature} select the optimal bands by using an exhaustive search in such a way that the bands are first evaluated individually for face recognition, and a combination of the results are then selected by using a score-level fusion method. However, evaluating  each band individually may not consider the complementary relationships between different bands. As a result, the selected subset of bands may not provide an optimal solution. To address this problem, Uzair \textit{et al.} \cite{uzair2015hyperspectral} leverage a sequential backward selection algorithm to search for a set of most discriminative bands. Sharma \textit{et al.} \cite{sharma2016hyperspectral} adopt a CNN-based model for band selection which uses  a  CNN to obtain the features from each spectral band independently, and then they use Adaboost in a greedy fashion (similar to other methods in the literature) for feature selection to determine the best bands. This method selects one band at a time, which ignores the complementary relationships between different bands for face recognition.

 In this book chapter, we propose a CNN-based model which adopts a Structural Sparsity Learning (SSL) technique to select the optimal bands to obtain the best recognition performance over \textit{all}  broad band images. We employ a group Lasso regularization algorithm \cite{yuan2006model} to sparsify the redundant spectral bands for face recognition. The group Lasso  puts a constraint on the structure of the filters in the first layer of our CNN during the training process. This constraint is a loss term augmented to the total loss function used for face recognition to zero out the redundant bands during the training of the CNN.
 To summarize, the main contributions of this book chapter include: 

{1: Joint face recognition and spectral band selection}: We propose an end-to-end deep framework which jointly recognizes hyperspectral face images and selects the optimal spectral bands for the face recognition task.
 
{2: Using group sparsity to automatically select the optimal bands}: We adopt a group sparsity technique to reduce the depth of convolutional filters in the first layer of our CNN network. This is done to zero out the redundant bands during face recognition. Contrary to most of the existing methods which select the optimal bands in a greedy fashion or manually, our group sparsity technique selects the optimal bands automatically to obtain the best face recognition performance over all the spectral bands. 
 
{3: Comprehensive evaluation and obtaining the best recognition accuracy}:  We evaluate our algorithm comprehensively on three standard publicly available hyperspectral face image datasets. The results indicate that our method outperforms state of the art spectral band selection methods for face recognition.
 
 \section{Sparsity}
  The Sparsity of signals has been a powerful tool
in many classical signal processing applications, such as
denoising and compression. This is because most natural signals can be represented compactly by only a few coefficients that carry
the most principal information in a certain dictionary or basis.
Currently, applications in sparse data representation have also been leveraged to the field of  pattern recognition and computer vision by the  development of  compressed sensing (CS)
framework and sparse modeling of signals and images. These applications are essentially based on the fact that, when
contrasted to the high dimensionality of natural signals, the signals in the same category usually exist in a low-dimensional subspace. Thus, for each sample, there is a sparse representation with respect to some proper basis which encodes
the important information. The CS concepts guarantee that a sparse signal can be recovered from its incomplete but incoherent projections with a high probability. This enables the recovery of the sparse representation by decomposing the sample over an often over-complete dictionary constructed by or learned from the representative samples. Once the sparse representation vector is constructed, the important information can be obtained  directly from the recovered vector.

 Sparsity was also introduced to enhance the accuracy of prediction and interpretability of regression models by altering the model fitting process to choose only a subset of  provided covariates for use in the final model rather than using all of them. Sparsity is important for many reasons as follows:

a) It is essential to havesas the smallest  possible number of neurons in neural network firing at a given time when a stimulus is presented. This means that a sparse model is faster as it is possible to make use of that sparsity to construct faster specialized algorithms. For instance, in structure from motion, the obtained data matrix is sparse when applying bundle adjustments of many methods that have been proposed to take advantage of the sparseness and  speedup things.
Sparse models are normally very scalable but they are compact. Recently, large scale deep learning models can easily have larger than 200k nodes. But why are they not very functional? This is  because they are not sparse.

b) Sparse models can allow more functionalities to be compressed into a neural network. Therefore, it is essential to have sparsity at the neural activity level in deep learning and exploring a way to keep more neurons inactive at any given time through neural region specialization. Neurological studies of biological brains indicate this region specialization is similar to face regions firing if a face is presented, while other regions remain mainly inactive. This means finding ways to channel the stimuli to the right regions of the deep model and prevent computations that end up resulting in no response. This  can help in making deep model not only more efficient but more functional as well.

c) In a deep neural network architecture, the main characteristic that matters is sparsity of connections; each unit should often be connected to comparatively few other units. In the human brain, estimates of the number of neurons are around $10^{10}$-$10^{11}$ neurons. However, each neuron is only connected to about $10^4$ other neurons on average. In deep learning, we see this in convolutional networks architectures. Each neuron receives information only from a very small patch in the lower layer.

d) Sparsity of connections can be considered as resembling sparsity of weights. This is because it is equivalent to having a fully connected network that has zero weights in most places. However, sparsity of connections is better, because we do not  spend the computational cost of explicitly multiplying each input by zero and augmenting all those zeros.

 Statisticians usually learn sparse models  to understand which variables are most critical.  However, it is an analysis strategy, not a strategy for making better predictions. The process of learning activations that are sparse does not really seem to matter as well. Previously, researchers thought that part of the reason that the Rectified Linear Unit (ReLU) worked well was that they were sparse. However, it was shown that all that matters is that they are piece-wise linear.

 \section{Compression approaches for neural networks }
 
 Our algorithm is closely related to a compression technique based on sparsity. Here, we also provide a brief
overview of other two popular methods: quantization and decomposition. 
\subsection{Network pruning}
Initial research on neural network compression concentrates on removing useless
connections by  using weight decay. Hanson and Pratt \cite{hanson1989comparing} propose hyperbolic and exponential
biases to the cost objective function. Optimal Brain Damage and Optimal Brain Surgeon \cite{hassibi1993second} prune the networks by using second-order derivatives of the objectives.
Recent research by Han \textit{et al.} \cite{han2015learning} alternates between pruning near-zero weights, which are
encouraged by $\ell_1$ or $\ell_2$ regularization, and retraining the pruned networks.
More complex regularizers have also been introduced. Wen \textit{et al.} \cite{wen2016learning} and Li \textit{et al.} \cite{li2016pruning} place
structured sparsity regularizers on the weights, while Murray and Chiang \cite{murray2015auto} place them on the hidden
units. Feng and Darrell \cite{feng2015learning} propose a nonparametric prior based on the Indian buffet processes
\cite{griffiths2011indian} on the network layers. Hu \textit{et al.} \cite{hu2016network} prune neurons based on the analysis of
their outputs on a large dataset. Anwar \textit{et al.} \cite{anwar2017structured} use particular sparsity patterns: channel-wise
(deleting a channel from a layer or feature map), kernel-wise (deleting all connections between two
feature maps in successive layers), and intra-kernel-strided (deleting connections between two
features with special stride and offset). They also introduce the use of a particle filter to point out the
necessity of the connections and paths over the course of training.
Another line of research introduces fixed network architectures with some subsets of connections deleted.
For instance, LeCun \textit{et al.} \cite{lecun1990optimal} delete connections between the first two convolutional
feature maps in an entirely uniform fashion. This approach, however, only considers a pre-defined pattern in which the same number of input feature map are assigned to each
output feature map. Moreover,
this method does not investigates how sparse connections influence the performance compared to dense networks.

Likewise, Cireşan \textit{et al.}. \cite{cirecsan2011high}  delete random connections in their MNIST experiments.
However, they do not aim to preserve the spatial convolutional density and it may be challenging to harvest the savings on existing hardware. Ioannou \textit{et al.}  \cite{ioannou2017deep} investigate three kinds of
hierarchical arrangements of filter groups for CNNs, which depend on different assumptions about
co-dependency of filters within each layer. These arrangements contain columnar topologies which are inspired
by AlexNet \cite{krizhevsky_imagenet_2012}, tree-like topologies have been previously used by Ioannou \textit{et al.} \cite{ioannou2017deep}, and root-like topologies. Finally, \cite{howard2017mobilenets} introduces the depth multiplier technique to
scale down the number of filters in each convolutional layer by using a scalar. In this case, the depth multiplier
can be considered as a channel-wise pruning method, which has been introduced in \cite{anwar2017structured}. However, the depth
multiplier changes the network architectures before the training phase and deletes feature maps of each layer
in a uniform fashion.
With the exception of \cite{anwar2017structured} and the depth multiplier \cite{howard2017mobilenets}, the above previous work performs connection pruning that causes nonuniform
network architectures. Therefore, these approaches need additional efforts to represent network connections
and may or may not lead to a reduction in  computational cost.

\subsection{Quantization}

Decreasing the degree of redundancy of the parameters of the model can be performed in the form
of quantization of the network parameters. Arora \textit{et al.} \cite{arora2014provable} propose to train CNNs with binary and ternary weights, accordingly. Gong \textit{et al.} \cite{gong2014compressing} leverage vector quantization for parameters in fully connected
layers. Anwar \textit{et al.} \cite{anwar2015fixed} quantize a network with the squared error minimization. Chen \textit{et al.} \cite{chen2015compressing} group network parameters randomly by using a hash function. Note that this method can
be complementary to the network pruning method. For instance, Han \textit{et al.} \cite{han2016dsd} merge connection pruning
in (Han \textit{et al.} \cite{han2015learning}) with quantization and Huffman coding.

\subsection{Decomposition}
Decomposition is another method which is based on low-rank decomposition of the parameters. Decomposition
approaches include truncated Singular Value Decomposition (SVD) \cite{denton2014exploiting}, decomposition to rank-1 bases \cite{jaderberg2014speeding}, Canonical Polyadic Decomposition (CPD) \cite{lebedev2014speeding},
 sparse dictionary learning , asymmetric (3D) decomposition by using reconstruction
loss of non-linear responses which is integrated with a rank selection method based on Principal Component Analysis (PCA)  \cite{zhang2016accelerating}, and Tucker decomposition by  applying a kernel tensor reconstruction loss
which is integrated with a rank selection approach based on global analytic variational Bayesian matrix factorization \cite{kim2015compression}.

\section{Regularization of neural network}

Alex \textit{et al.} \cite{krizhevsky_imagenet_2012} proposed Dropout to regularize fully connected layers
in the neural networks layers by randomly setting a subset of activations to zero over the course of training. Later, 
Wan \textit{et al.} \cite{wan2013regularization} introduced DropConnect, a generalization of Dropout that instead randomly zero out
a subset of weights or connections. Recently, Han \textit{et al.} \cite{han2016dsd} and Jin \textit{et al.} \cite{jin2016training} propose a kind of regularization where dropped connections are unfrozen and the network is retrained. This method can be thought of as an incremental training
approach. 
\section{ Neural network architectures}
Network architectures and compression   are closely related. The purpose of compression is to eliminate
redundancy in network parameters. Therefore, the knowledge about traits that indicate the success of architecture
success is advantageous. Other than the discovery that depth is an essential factor, little is  known regarding such traits. Some previous research performs architecture search but without the main purpose of performing compression. Recent work introduces skip connections or shortcut 
to convolutional networks such as 
residual networks \cite{he_resnet_2016}.

\section{Convolutional Neural Network}
CNN is a well-known used deep learning framework which was inspired by the visual cortex of animals. First, it was widely applied for object recognition but now it is used in other areas as well like object tracking \cite{fan2010human}, pose estimation \cite{toshev2014deeppose}, visual saliency detection \cite{zhao2015saliency}, action recognition \cite{donahue2014decaf}, and object detection \cite{zhao2019object}.  CNNs are similar to traditional neural network in such a way that they are consists of neurons that self-optimize through learning. Each
neuron receives an input and conduct an operation (such as a 
product of scalar followed by a non-linear function) the basis of countless neural networks. From the given input image  to the final output of the class score, the entire of the network still represents a single perceptive score function.
The last layer  consists of a loss functions associated with the classes, and all of the regular methodologies and  techniques introduced for traditional neural network still can be used.
The only important difference between CNNs and traditional neural network is that CNNs are essentially used in the field of pattern recognition within images. This gives us the opportunity 
 to encode image-specific features into the architecture, making the network more suited for image-focused tasks,  while further reducing the parameters
required to set up the model.
One of the largest limitations of traditional forms of neural network is that they aim to
challenge with the computational complexity needed to compute image data.
Common machine learning  datasets such as the MNIST database
of handwritten digits are appropriate for most types of neural network, because of its relatively
small image dimensionality of just $28 \times 28$. With this dataset, a single neuron in
the first hidden layer will consists of 784 weights ($28 \times 28 \times 1$ where one considers
that MNIST is normalized to just black and white values), which can be controlled 
for most types of neural networks. Here, we used a CNN for our hyperspectral band selection for face recognition. We used the VGG-19 \cite{simonyan_very_deep_2014} as our  baseline – CNN. \cite{simonyan2014very}.
\subsection{Convolutional Layer}
\begin{figure}[t]
\centering
\includegraphics[scale=0.3]{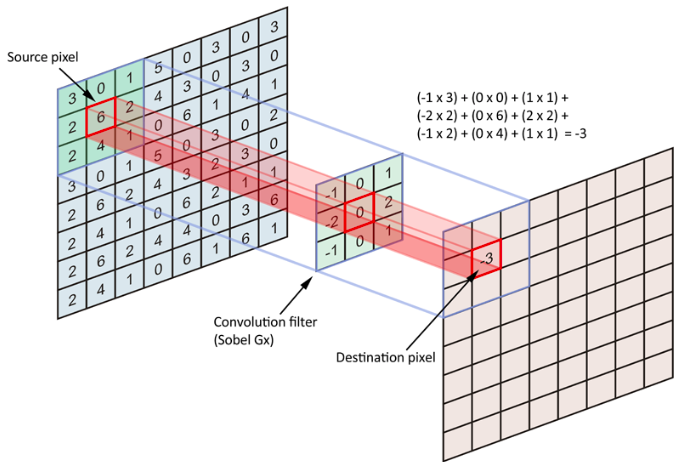}
\caption{Convolution operation.}
\label{hir}
\end{figure}

The convolutional layer constructs the basic unit of a CNN where most of the computation is conducted. It is basically a set of feature maps with neurons organized in it. The weights of the convolutional layer are a set of  filters or kernels which are learned during the training. These filters are convolved by the feature maps to create a separate two-dimensional activation map  stacked together alongside the depth dimension, providing the output volume. Neurons that exist in the same feature map shares the weight whereby decreasing the complexity of the network by keeping the number of weights low. The spatial extension of sparse connectivity between the neurons of two layers is a hyperparamter named the receptive field. The hyperparameters that manage the size of the output volume are the depth (number of filters at a layer), stride (for moving the filter) and zero-padding (to manage spatial size of the output). The CNNs are trained by back-propagation and the backward pass as well, performs a convolution operation, but with spatially flipped filters. Fig. \ref{hir} shows the basic convolution operation of a CNN.

One of the  traditional versions of a CNN is "Network In Network"(NIN), introduced by Lin \textit{et al.} \cite{zeiler2014visualizing}, where the $1\times1$ convolution filter leveraged is a Multi-Layer Perceptron (MLP) instead of the typical linear filters and the fully connected layers are replaced by a Global Average Pooling (GAP) layer. The output structure is named the  MLP-Conv layer because the micro network contains of stack of MLP-Conv layers. Dissimilar to a regular CNN, NIN can improve the abstraction ability of the latent concepts. They work very well in providing for justification that the last MLP-Conv layer of NIN were confidence maps of the classes  leading to the possibility of conducting object recognition using  NIN. The GAP layer within  the architecture is used to reduce the parameters of our framework. Indeed, reducing the dimension of the CNN output by the GAP layer prevents our model from becoming over-parametrized  and  having a large dimension. Therefore,  the chance of overfitting in model is potentially reduced.

\subsection{Pooling Layer}

\begin{figure}[t]
\centering
\includegraphics[scale=1]{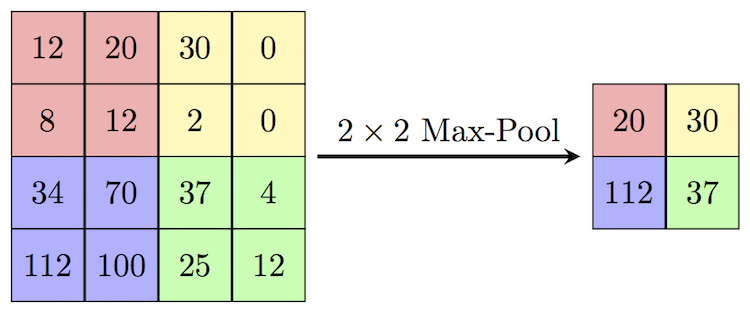}
\caption{Max pooling operation.}
\label{hir1}
\end{figure}

Basic CNN architectures have alternating convolutional and pooling layers and the latter functions to reduce the spatial dimension of the activation maps (without loss of information) and the number of parameters in the network and therefore decreasing  the overall computational complexity. This manages the problem of overfitting. Some of the common pooling operations are max pooling, average pooling, stochastic pooling \cite{zeiler2013stochastic}, spectral pooling \cite{rippel2015spectral}, spatial pyramid pooling \cite{nguyen2015deep} and multiscale orderless pooling \cite{gong2014multi}. The work by Alexey Dosovitskiy \textit{et al.} \cite{springenberg2014striving} evaluates the functionality of different components of a CNN, and has found that max pooling layers can be replaced with convolutional layers with stride of two. This  essentially can be applied for simple networks which have proven to beat many existing intricate  architectures. We used max-pooling in our deep model. Fig. \ref{hir1} shows the operation of max pooling.

\subsection{Fully Connected Layer}

Neurons in this layer are Fully Connected (FC) to all neurons in the previous layer, as in a regular neural network. High level reasoning is performed here. The neurons are not spatially arranged  so there cannot be a convolution layer after a fully connected layer. Currently, some deep architecture have their FC layer replaced, as in NIN, where FC layer is replaced by a GAP layer. 

\subsection{Classification Layer}
The last FC layer serves as the classification layer that calculates the loss or error which is a penalty for discrepancy between actual output and desired. For predicting a single class out of $k$ mutually exclusive classes, we use Softmax loss. It is the commonly and widely used loss function. Specifically, it is multinomial logistic regression. It maps the predictions to non-negative values and is normalized to achieve probability distribution over classes. Large margin classifier, SVM, is trained by computing a Hinge loss. For regressing to real-valued labels, Euclidean loss can be calculated. We used Softmax loss to train our deep model. The Softmax loss is formulated as follows:
\begin{equation}
  \mathcal{L}({w})= -\sum_{i=1}^n \sum_{j=1}^k y_i^{(j)} log ( p_i^{(j)}),
\end{equation} where, $n$ is the number of training samples, $y^{(i)}$ is the one-hot encoding label for the $i$-th sample, and $y_i^{(j)}$ is the $j$-th element in the label vector $y_i$. The varaible $p_i$ is the probability vector  and $p_i^{(j)}$ is the $j$-th element in the label vector $p_i$ which indicate the probability that CNN assign to class $j$.  The varaible $w$ is  the parameter of the CNN.
 \begin{figure}[t]
\centering
\includegraphics[scale=0.3]{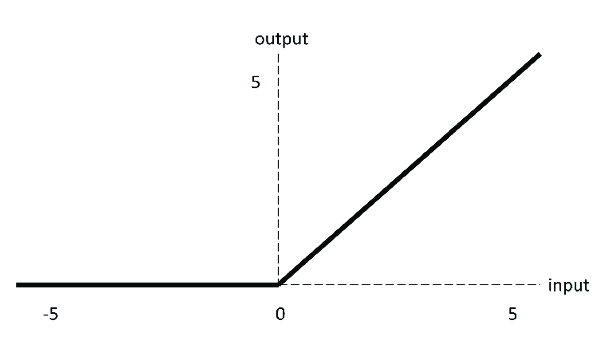}
\caption{ReLu activation function.}
\label{rel}
\end{figure}
\subsection{Activation Function: ReLU}
ReLU is the regular activation function that is used in CNN models. It is a linear activation function which has thresholding at zero as shown in Eq.2. It has been shown that the convergence of gradient descent is accelerated by applying ReLU. The
ReLU activation function is shown in Fig. \ref{rel} 
\begin{equation}
f(x)= max \{0, x\}.
\end{equation}
\subsection{VGG-19 Architecture}
\begin{figure*}[t]
\centering
\includegraphics[scale=0.3]{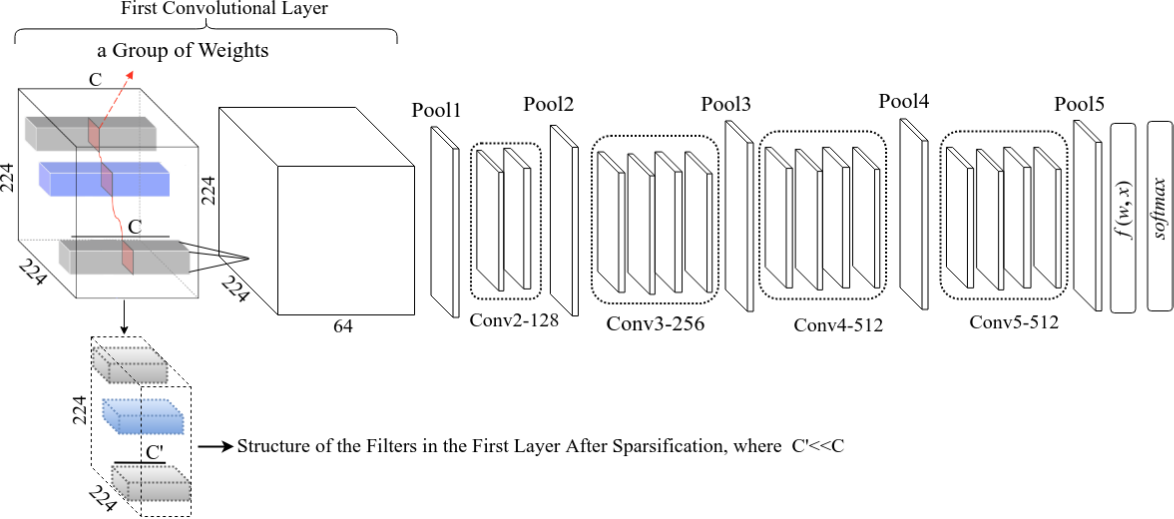}
\caption{Block diagram of hyperspectral band selection for face recognition based on structurally sparsified CNN.}\label{arch}
\end{figure*}

Our band selection algorithm can be used  for any other deep architecture including ResNet \cite{zagoruyko2016wide}, and AlexNet \cite{krizhevsky2012imagenet},  and there is no restriction on choosing a specific deep model during the process of band selection in the first convolutional layer of these networks using our algorithm. We used VGG-19 network since a) it is easy to implement in Tensorflow and it is more popular than other deep models and b) it achieved excellent results on the ILSVRC-2014  dataset (i.e., ImageNet competition). The input to our VGG-19 based CNN is a fixed-size $224 \times 224 $ hypespectral image. The only pre-processing that we perform  is to subtract the mean spectral value, calculated on the training set, from each pixel. The image is sent through a stack of convolutional operation, where we use filters with a very small receptive field of $3 \times 3$. This filter size is the smallest size that capture the notion of left and right, up and down, and center. In one of the configurations,  we also can  use  $1 \times 1$ convolutional filters, which can be considered as a linear transformation of the input channels. The convolutional stride is set to 1 pixel. The spatial padding of the convolutional layer input is such that the spatial resolution is preserved after convolution, which means that the padding is 1 pixel for $3 \times 3$ convolutional layers. 
Spatial pooling is performed by five max-pooling layers, which follow some of the convolutional layers. Note that not all of the convolutional layers are followed by max-pooling. In VGG-19 network, max-pooling is carried out on a $2 \times 2$ pixel window, with stride of 2. A stack of convolutional layers is followed by two FC layers as follows: the first  has 4096 nodes, the second performs $k$ nodes (i.e., one for each class). The second layer is basically the soft-max layer. The hidden layer is followed by rectification ReLU non-linearity. The overall architecture of VGG-19 is shown in Fig. \ref{arch}.

\section{SSL Framework for Band Selection}
 We propose a regularization  scheme which uses a SSL  technique  to  specify the optimal  spectral  bands  to  obtain  the best  face  recognition  performance over all the spectral bands. Our regularization method is based on a  group Lasso algorithm \cite{yuan2006model} which shrinks a set of groups of weights during the training of our CNN architecture.  By using  this regularization method, our algorithm recognizes face images with  high accuracy, and simultaneously, forces  some  groups of weights corresponding to redundant bands to become zero. In our framework, the goal is achieved by adding the $ \ell _{12}$ norm of the groups as a sparsity constraint term to the total loss function of the network for face recognition. Depending on how much sparsity that we want to impose to our model, we scale the sparsity term by a hyperparameter. The hyperparameter creates a balance between face recognition loss  and the sparsity constraint during the training step. It can be shown that if we enlarge the hyperparameter value, we impose more sparsity on our model, and if the hyperparameter is set to a value close to  zero, we add less sparsity constraint to our model.
\subsection{Proposed Structured Sparsity learning for generic structures}

 In our regularization framework, the hyperspectral images are directly fed to the CNN. Therefore, the depth of each convolutional filter in the first layer of the CNN is equal to the number of spectral bands, and all the weights belonging to the same channel for all the convolutional filters in the first layer construct a group of weights. This results in the number of groups in our regularization scheme being equal to the number of  spectral bands. The group Lasso regularization algorithm attempts to zero out the groups of  weights that are related to the redundant bands during the training of our CNN.
 \subsection{Total Loss Function of the Framework}
 Suppose that $w$ is all the weights for the convolutional filters of our CNN, and $w_1$ denotes  all the weights in the first convolutional layer of our CNN. Therefore, each weight in a given layer is identified by a 4-D tensor (i.e., ${\rm I\!R} ^ {L \times C \times P \times Q}$, where $L$, $C$, $P$, and $Q$ are the dimensions of the weight in the tensor space along the axes of  the filter, channel,  spatial height, and width, respectively).
 The proposed  loss function which uses SSL to train our CNN is formulated as follows:
\begin{equation}
     \mathcal{L}({w})= \mathcal{L}_{r}(w)+ \lambda_g . \mathcal{R}_g({w}_{1}),
\end{equation}
where $\mathcal{L}_r(.)$ is loss function used for  face recognition, and $\mathcal{R}_g(\text{.})$ is SSL loss term  applied  on the convolutional filters in the first layer. The variable $\lambda_g$ is a hyperparameter used to balance the two loss terms in (3). Since group Lasso can effectively zero out all of the weights in some groups \cite{yuan2006model}, we leverage it in our total loss function to zero out groups of  weights corresponding to the redundant spectral bands in the band selection process. Indeed the total loss function in (3) consists of two terms in which the first term performs face recognition, while the second term performs band selection based on the SSL. These two terms are optimized jointly during the training of the network. 
\subsection{Face Recognition Loss Function} 
 In this section, we describe the loss function, $\mathcal{L}_r(w)$, that we have  used for face recognition. We use the center loss \cite{wen2016discriminative} to learn a set of  discriminative features for hyperspectral face images. The softmax classifier loss is typically used in a CNN only forces the CNN features of different classes to stay apart. However, the center loss not only does this, but also efficiently brings the CNN features of the same class close to each other. Therefore, by considering the center loss during the training of the network, not only are the inter-class feature differences enlarged, but also the intra-class feature variations are reduced. The center loss function for face recognition is formulated as follows:
 \begin{equation}
  \mathcal{L}_{r}({w})= -\sum_{i=1}^n \sum_{j=1}^k y_i^{(j)} log ( p_i^{(j)})+\frac{\gamma}{2} \sum_{i=1}^n ||f({w},x_i)-{c}_{yi}||_2^2,
\end{equation}
 where $n$ is the number of  training data, $ f({w},x_i)$ is the output of the CNN, $x_i$ is the $i^{\text{th}}$ image in the training batch. The variable $y_i$ is one hot encoding label corresponding to the sample $x_i$, and $y_i^{(j)}$ is the $j^{\text{th}}$ element in  vector $y_i$, $k$ is the number of classes, and $p_i$ is the output of the softmax applied only on the output of the CNN (i.e., $ f({w},x_i)$). The variable ${c}_{yi}$ indicates  the center of  the features corresponding to the $i^ {\text{th}}$ class. The variable $\gamma$ is  a hyperparameter used to balance the two  terms in the center loss.
 \subsection{ Band Selection via Group Lasso}
Assume that each hyperspectral image has $C$ number of spectral bands. Since, in our regularization scheme,  hyperspectral images are directly fed to the CNN, the depth of each convolutional filter in the first layer of our CNN is equal to $C$. Here, we adopt a group Lasso to regularize the depth of each convolutional filter in the first layer of our CNN. We use the group Lasso because it can effectively zero out all of the weights in some groups \cite{yuan2006model}. Therefore, the group Lasso can zero out groups of  weights which correspond to redundant spectral bands. In the setup of our group Lasso regularization, weights  belonging to the same channel for all the convolutional filters in the first layer form a group (red squares in Fig. \ref{arch})  which can be removed  during the training step by using $\mathcal{R}_g({w}_{1})$ function as defined in (3). Therefore, there are $C$ number of groups in our regularization framework.
The group Lasso regularization on the parameters of ${w}_{1}$ is an $\ell_{12}$ norm which can be expressed as follows:
\begin{equation}
    \mathcal{R}_g({w}_{1})=\sum_{g=1}^C ||{w}_{1}^{(g)}||_2,
\end{equation}
 where ${w}_{1}^{(g)}$ is the subset of weights (i.e., a group of weights) from ${w}_{1}$, 
and $C$ is the total number of groups. Generally, different groups may overlap in the group Lasso regularization. However, this does not happen in our case. The notation $|| \textbf{.} ||_2$ represents an  $\ell_{2}$ norm on the parameters of the group ${w}_{1}^{(g)}$.
 Therefore, the group Lasso regularization as a sparsity constraint for band selection can be expressed as follows:
\begin{equation}
R_g({w}_{1})=\sum_{c=1}^C \sqrt{\sum_{l=1}^L \sum_{p=1}^P\sum_{q=1}^Q ({w}_{1}(l,c,p,q))^2},
\end{equation}
where ${w}_{1}(l,c,p,q)$ denotes a weight located in $l^\text{\ th}$ convolutional filter, $c^\text{\ th}$ channel, and $(p,q)$ spatial position.  In this formulation, all of the weights $w_1(:,c,:,:)$ (i.e., the weights which have the same index $c$), belong to the same group ${w}_{1}^{(c)}$. Therefore, $R_g({w}_{1})$ is an $\ell_{12}$ regularization term in which $\ell_1$ is performed on the $\ell_2$ norm of each group.
\subsection{Sparsification Procedure}
The proposed framework automatically selects the optimal bands from all spectral bands for face recognition during the training phase.
For clarification, we can assume that in a typical RGB image, we  have three  bands and  the depth of each filter in the first convolutional layer is three.  However, here, there are $C$ spectral bands and as a consequence, the depth of each filter in the first layer is $C$. As shown in Fig. \ref{arch}, hyperspectral images are fed into the CNN directly. The group Lasso efficiently removes  redundant  weight groups (associated with different spectral band)  to improve the recognition accuracy during the training phase. In the beginning of the training, the depth of the filters is $C$, and once we start to sparsify the depth of the convolutional filters, the depth of  each filter will be reduced (i.e., $C' << C$). 

It should be noted that the dashed cube in the Fig. \ref{arch} is not part of our CNN architecture. This is the structure of the  convolutional filters in the first layer after several epochs training the network using the network loss function defined in (3). 
\section{Experimental Setup and Results}
 \begin{figure*}
\centering
\includegraphics[trim={8cm 0cm 3cm 1cm}, scale=0.33]{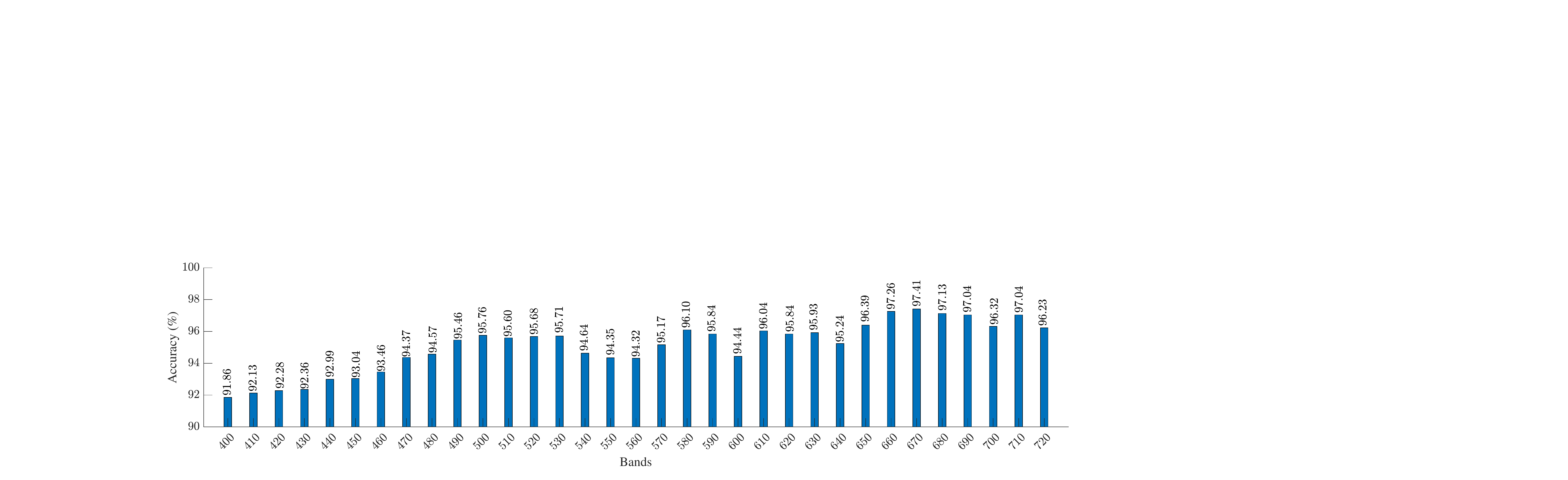}
\caption{Face recognition accuracy of each individual band on the UWA -HSFD}
\label{UWA}
\end{figure*}
\subsection{CNN  Architecture} We  use the  VGG-19 \cite{simonyan2014very}  architecture  as  shown  in  Fig. \ref{arch} with  the same  filter  size,  pooling  operation  and  convolutional  layers. However, the depth of the filters in the first convolutional layer of our CNN is set to the number of the hyperspectral bands. The network uses filters with a receptive field of $3\times3$. We set the convolution stride to 1 pixel. To preserve spatial resolution after convolution, the spatial padding of the convolutional layer is fixed to 1 pixel for all the $3\times3$ convolutional  layers.  In  this  framework,  each  hidden  layer  is  followed by a ReLU activation function. We apply batch normalization (i.e., shifting inputs to zero-mean and unit variance) after each convolutional and fully connected layer, and before performing the  ReLU  activation  function.  Batch normalization potentially helps to achieve faster learning as well as higher overall accuracy. Furthermore, batch normalization allows us to use a higher learning rate, which potentially provides another boost in speed.
\subsection{Initializing Parameters of the Network}
In this section, we describe how we initialize the parameters of our network for the training phase.
Thousands of images are needed to train such a deep model.
For this reason, we initialize the  parameters of our network by a VGG-19 network
pre-trained on the ImageNet dataset and then we fine tune
it as a classifier by using the CASIA-Web Face dataset \cite{yi2014learning}.
CASIA-Web Face contains 10,575 subjects and 494,414
images. As far as we know, this is the largest publicly available
face image dataset, second only to the private Facebook
dataset. In our case, however, since   the depth of the filters in the first layer is  the number of spectral bands, we initialize these filters  by duplicating the filters of  the pre-trained VGG-19 network in the first convolutional layer.  For example, assume that the depth of the filters in the first layer is $3n$ (we have $3n$ spectral bands). Then, in such a case, we duplicate filters of the first layer $n$ times as an initialization point for training the network.
\subsection{Training the Network}
We use the Adam optimizer \cite{kingma2014adam}  with the default hyper-parameter values ($\epsilon = 10^{-3}$, $\beta_1 = 0.9$, $\beta_2 = 0.999$) to minimize the total loss function of our network defined in (3). The Adam optimizer is a robust and
well-adapted optimizer that can be applied to a variety of
non-convex optimization problems in the field of deep neural
networks. We set the learning
rate to 0.001 to minimize loss function (3) during the training process. The hyperparameter $\lambda_g$ is selected by cross-validation in our experiments.  We ran the CNN model through 100 epochs, although the model nearly converged after 30 epochs. The batch size in all experiments is fixed to 32. We implemented our algorithm in TensorFlow, and all experiments are conducted on two GeForce GTX TITAN X 12GB GPUs.
\begin{figure*}
\centering
\subfigure[CMU-HSFD]{\includegraphics[scale=0.465]{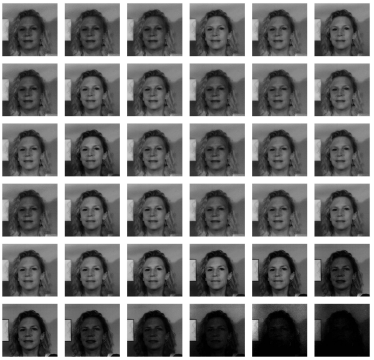}}
\subfigure[UWA-HSFD]{\includegraphics[scale=0.63]{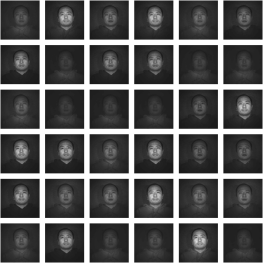}}
\subfigure[HK PolyU-HSFD]{\includegraphics[scale=0.8]{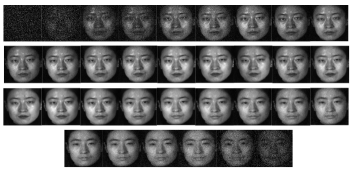}}
\caption{Samples of hyperspectral images.}
\label{Rank_LFW}
\end{figure*}
\subsection{Hyperspectral Face Datasets}
We performed our experiments on three standard and   publicly available hyperspectral face image datasets including CMU \cite{denes2002hyperspectral}, HK PolyU \cite{di2010studies}, and  UWA \cite{uzair2013hyperspectral}. Descriptions of these datasets are as follows:

\textbf{CMU-HSFD:} The face cubes in this dataset have been obtained
by a spectro-polarimetric camera. The spectral wavelength range during the image acquisition is from 450 $nm$ to
1100 $nm$ with a step size of 10 $nm$. The images of this dataset have been
collected in multiple sessions from 48 subjects. 

\textbf{HK PolyU-HSFD:}  The face
images in this dataset have been obtained by using an indoor system made up of CRI's VariSpec Liquid Crystal Tunable Filter with a halogen light source. The  spectral wavelength range  during  the image  acquisition  is from 400 $nm$ to 720 $nm$ with a step size of 10 $nm$,  which creates 33 bands in total. There are 300 hyperspectral face
cubes captured from 24 subjects. For each subject, the hyperspectral face cubes have been collected from multiple sessions in an average span of five months.

\textbf{UWA-HSFD:} Similar to the HK PolyU dataset, the face
images in this dataset have been acquired 
by using an indoor imaging system made up of CRI's VariSpec Liquid Crystal Tunable Filter integrated with a Photon focus camera. However, the camera exposure time is set  and altered based on the signal-to-noise ratio for
different bands. Therefore, this dataset has the advantage of having lower noise levels in comparison to other two
datasets. There are 70 subjects in this dataset, the  spectral wavelength range  during the image acquisition is from 400 $nm$ to 720 $nm$ with a step size of 10 $nm$. 

Table. \ref{tabb} indicates a summary of the datasets that we have used in our experiments. 
\begin{table}
    \centering
    \small
    \scalebox{1.2}{\begin{tabular}{c c c c c}
     Dataset &  Subjects  & HS Cubes & Bands &  Spectral Range \\
         \hline
    CMU &  48 & 147 & 65 & 450-1090 $nm$\\
    \hline
     HK PolyU & 24 & 113 & 33 &  400-720  $nm$\\
     \hline
     UWA & 70 & 120 & 33 & 400-720  $nm$\\
       
    \end{tabular}}
    \caption{A summary of Hyperspectral Face datasets}
    \label{tabb}
\end{table}
\begin{figure*}
\centering
\subfigure[UWA-HSFD]{\includegraphics[scale=0.75]{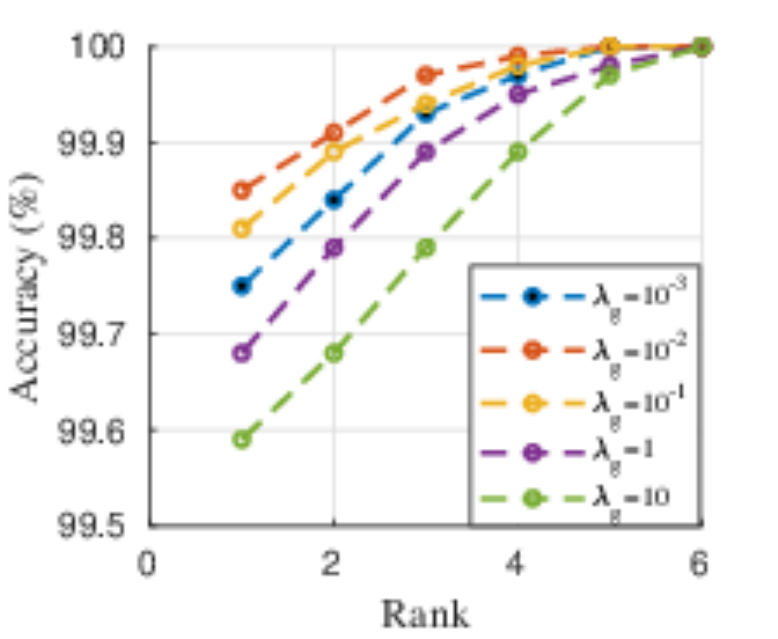}}
\subfigure[ HK PolyU-HSFD]{\includegraphics[scale=0.645]{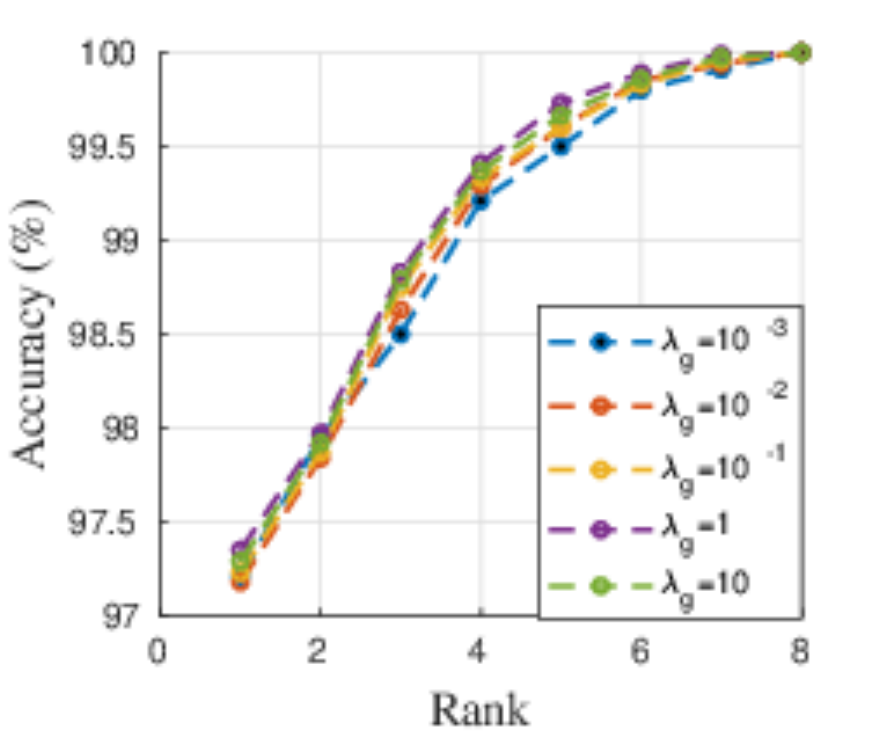}}
\subfigure[CMU-HSFD]{\includegraphics[scale=0.75]{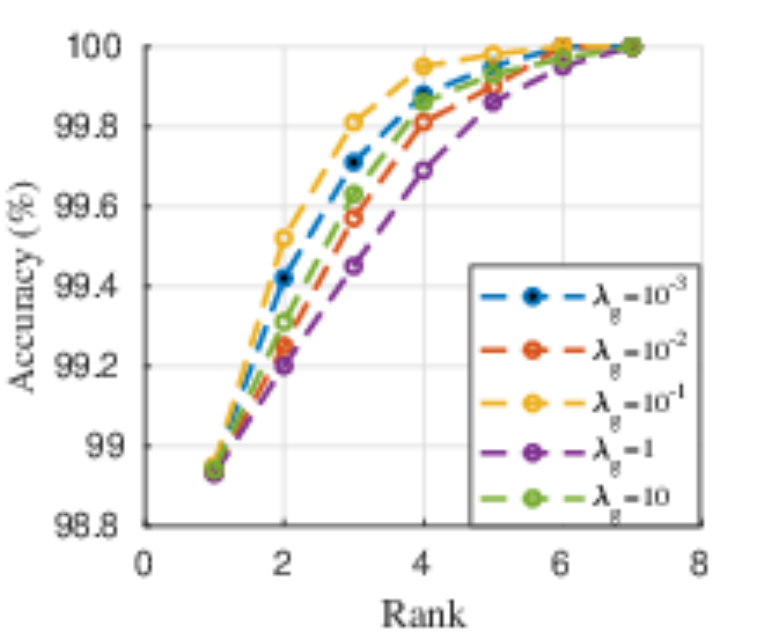}}
\caption{Accuracy of our model using different values of $\lambda_g$.}
\label{Rank_LFW}
\end{figure*}
\begin{figure*}
\centering
\includegraphics[trim={8cm 0cm 0cm 1cm}, scale=0.33]{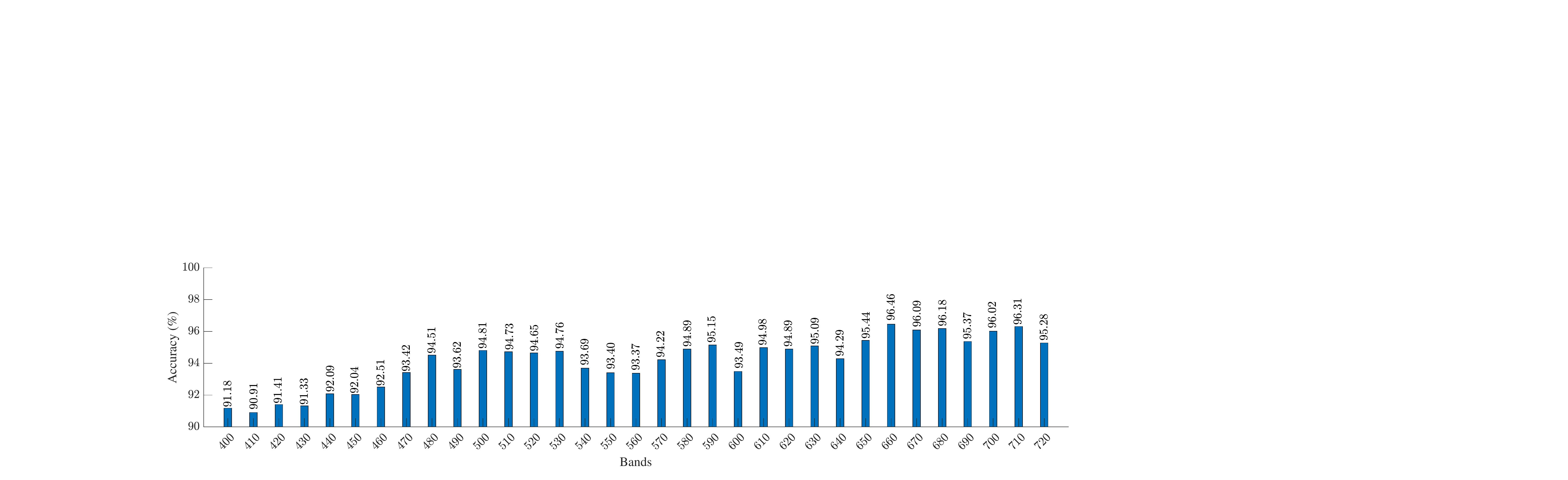}
\caption{Face recognition accuracy of each individual band on the HK PolyU-HSFD}
\label{HK}
\end{figure*}
\subsection{Parameter Sensitivity} 
We explore the influence of the hyper-parameter $\lambda_g$ defined in (3) on  face recognition performance. Fig. \ref{Rank_LFW} shows the CMC curves  for  CMU, HK PolyU, and UWA HSFD  with different values of $\{10, 1, 10^{-1},10^{-2}, 10^{-3}\}$, respectively. We can see that our network total loss   defined in  (3) is not significantly sensitive to $\lambda_g$ if we set these parameters within $[10^{-3},10]$ interval.
\subsection{Updating Centers in Center Loss}
We used the strategy presented in \cite{wen2016discriminative} to update the center of each class (i.e., ${c}_{yi}$ in (4)). In this strategy, first, instead of updating the centers with respect to the entire training set, we update the centers based on a mini-batch such that, in each iteration, the centers are obtained by averaging the features of the corresponding classes.  Second, to prevent the large perturbations made by a few mislabeled samples, we scale it by a small number of 0.001 to control the learning rate of the centers, as suggested in \cite{wen2016discriminative}.
 \begin{figure*}
\centering
\includegraphics[trim={8cm 0cm 1cm 0cm}, scale=0.246]{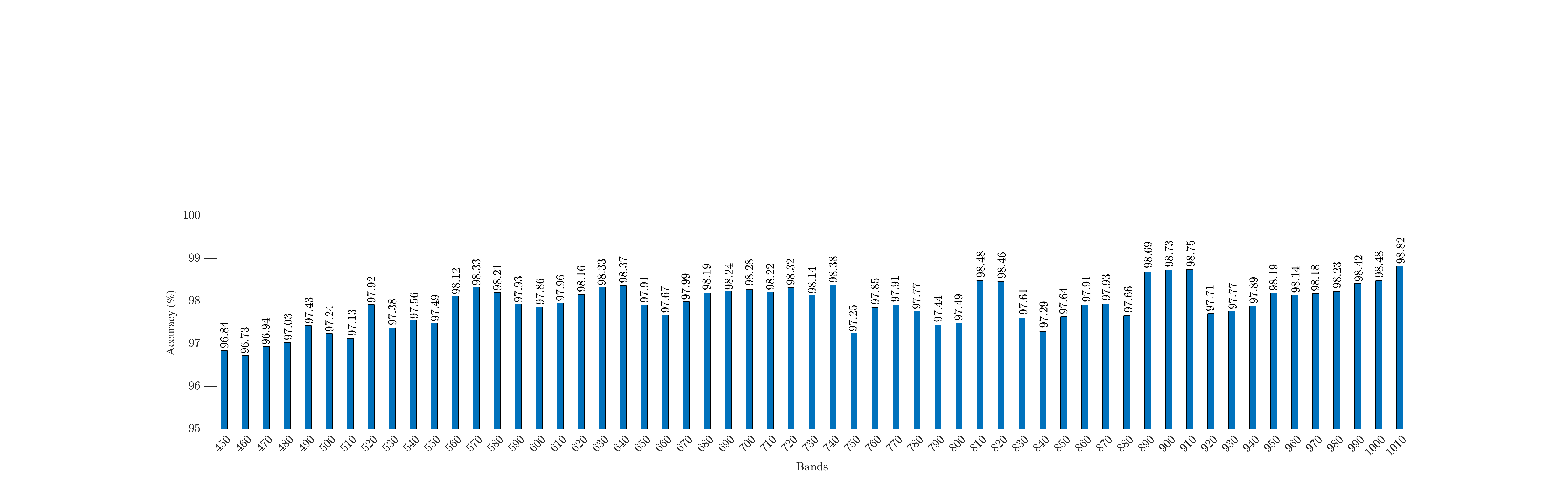}
\caption{Face recognition accuracy of each individual band on the CMU-HSFD}
\label{CMU}
\end{figure*}
\subsection{Band Selection}
RGB  cameras  produce 3 bands over the whole visible spectrum. However,   hyperspectral imaging  camera divides this range into many narrow bands (e.g., 10 nm).  Both of these types of imaging cameras are the extreme cases of spectral resolution. Even though RGB cameras divides the visible spectrum into three bands, they are  wide and the center of the wavelengths in these bands are selected to approximate the human visual system instead of maximizing the performance of the face recognition task. 

In this work, we conducted experiments to find the optimal number of bands and their center wavelengths that maximize  face recognition accuracy.  Our method adopts the SSL technique during the training of our CNN to automatically select  spectral bands which provide the maximum recognition accuracy. The results indicate that  maximum discrimination power can be achieved by using a small  number of bands rather than all the spectral bands but more than three bands in RGB for the CMU dataset. Specifically, the results demonstrate that the most discriminative spectral wavelengths for face recognition are obtained by  a subset of  red and  green wavelengths.

\begin{figure*}
\centering
\includegraphics[scale=0.3]{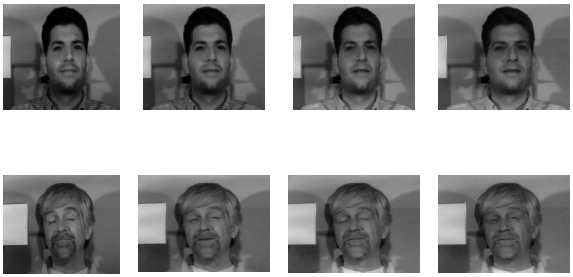}
\caption{Images of selected bands from CMU dataset.}
\label{UWAAA}
\end{figure*}

In addition to the improvement in face recognition accuracy, other advantage of the band selection include: a reduction in computational complexity, a reduction in the cost and time during image acquisition for hyperspectral cameras, and reduction in redundancy of the data. This is because one can capture the bands which are more discriminative for a face recognition task instead of capturing images from the entire visible spectrum.
\begin{figure*}
\centering
\includegraphics[scale=0.26]{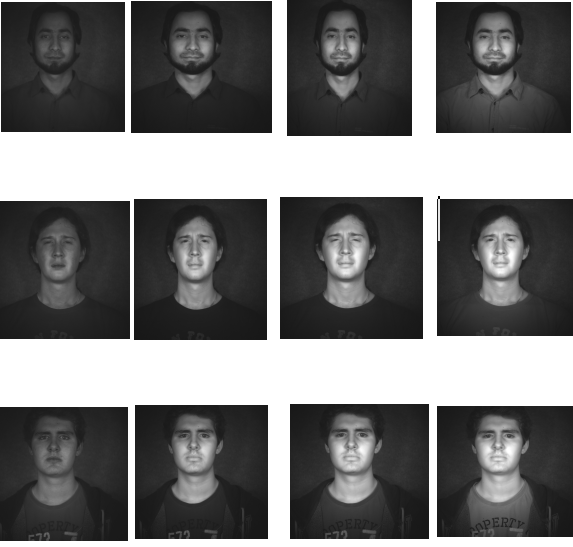}
\caption{Images of selected bands from UWA dataset.}
\label{UWAA}
\end{figure*}
Table. \ref{tab2} indicates the optimal spectral bands from all of the bands selected by our method. Our algorithm selects 4 bands including $\{ \text{750, 810 , 920, 990} \}$ for the  CMU dataset, 3 bands including $\{ \text{580, 640, 700} \}$ for PolyU, and 3 bands including $\{\text{570, 650, 680}\}$ for the UWA dataset. The results show that SSL selects the optimal bands from the green and red spectra and ignores bands within the blue spectrum. Fig. \ref{UWAA} and Fig. \ref{UWAAA} demonstrate some of the face images from the bands which are selected by our algorithm. The experimental results indicate that the blue wavelength bands are discarded earlier during the sparsification procedure  because they are less discriminative and they are less useful compared to the green, red and IR ranges for the task of face recognition. The group sparsity technique used in our algorithm automatically selects the optimal bands by combining the informative bands so that the selected bands have the most discriminative information for the task of face recognition.
\begin{table}
    \centering
    \small
    \scalebox{1.3}{\begin{tabular}{c c c c c}
     Dataset &    &   & Bands &  \\
         \hline
    CMU &   &  & \{750, 810, 920, 990 \}nm & \\
    \hline
     HK PolyU &  &  & \{580, 640, 700 \}nm & \\
     \hline
     UWA &  &  & \{ 570, 650, 680,710\}nm & \\
       
    \end{tabular}}
    \caption{Center wavelengths of the selected bands for different hyperspectral datasets.}
    \label{tab2}
\end{table}
\subsection{Effectiveness of SSL}
Fig. \ref{UWA}, Fig. \ref{HK}, and Fig. \ref{CMU}  indicate the face recognition accuracy for each individual  band on the UWA, CMU, and PolyU datasets, respectively. In Table. \ref{tab3}, we reported the maximum and minimum accuracy obtained from each spectral band when we use each band individually during the training. We also reported the case where we use all  bands without using the SSL technique for face recognition. Finally, we provided the results of our framework in the case where we use SSL during the training. The results show that using SSL not only removes the redundant spectral bands for the face recognition task, but  it can also improve the recognition performance in comparison to the case where all the spectral bands are used for face recognition. These improvements are around 0.59 \%, 0.36\%,and 0.32\% on the CMU, HK PolyU, and UWA datasets, respectively.
\begin{table}
    \centering
    \small
    \scalebox{1.3}{\begin{tabular}{c c c c c}
     Dataset &  Min  & Max & All the bands &  SSL \\
         \hline
    CMU &  96.73 & 98.82 & 99.34 & 99.93\\
    \hline
     HK PolyU & 90.91 & 96.46 & 99.52 & 99.88\\
     \hline
     UWA & 91.86 & 97.41 & 99.63 & 99.95\\
       
    \end{tabular}}
    \caption{Accuracy (\%) of  our band selection algorithm in different cases}
    \label{tab3}
\end{table}
\begin{table}
    \centering
    \small
    \scalebox{1.3}{\begin{tabular}{|c |c| c| c|}
     \hline
    \textbf{Methods}  &  \textbf{CMU}  & \textbf{PolyU} & \textbf{UWA}  
     \\
    \hline
   \textbf{Hyperspectral}  
      \\
    \hline
    Spectral Angle \cite{robila2008toward} & $38.1$ & $25.4$ & $37.9$
    \\
    Spectral Eigeface \cite{pan2009comparison} & $84.5$ &  $70.3$  & $91.5$
    \\
    2D PCA \cite{di2010studies}   & $72.1$  & $71.1$ & $83.8$
    \\
    3D Gabor Wavelets \cite{shen2012hyperspectral}  & $91.6$ & $90.1$ & $91.5$
     \\
    \hline
    \textbf{Image Set Classification}  
     \\
    \hline
    DCC \cite{kim2007discriminative}  & $87.5$ & $76.0$  & $91.5$
    \\
    MMD \cite{wang2008manifold}  & $90.0$ & $83.8$ & $82.8$
    \\
    MDA \cite{wang2009manifold}  &  $90.6$ & $87.9$  & $91.0$
    \\
    AHISD \cite{cevikalp2010face} & $90.6$ & $89.9$  & $92.5$
    \\
    SHIDS \cite{cevikalp2010face} & $91.1$ & $90.3$ & $92.5$
    \\
    SANP \cite{hu2012face}  & $90.9$ & $90.5$ & $92.5$
    \\
    CDL\cite{wang2012covariance}  & $92.7$ & $89.3$ & $93.1$
    \\
    PLS  \cite{uzair2015hyperspectral} & $99.1$ & $95.2$ & $98.2$
    \\
    PLS* \cite{uzair2015hyperspectral} & $99.1$ & $95.2$ & $98.2$
    \\
     \hline
    \textbf{Grayscale and RGB}
      \\
    \hline
    SRC \cite{wright2009robust}  & $91.0 $ & $85.6$ & $96.2$
    \\
    CRC \cite{zhang2011sparse}  & $93.8$ & $86.1$& $96.2$
    \\
    LCVBP+RLDA \cite{lee2012local} & $87.3$  & $80.3$ & $97.0$
    \\  
    \hline
    \textbf{CNN-Based Models}
      \\
    \hline
     S-CNN \cite{sharma2016hyperspectral} &  98.8 &  97.2 & -
    \\
     S-CNN+SVM \cite{sharma2016hyperspectral}  & 99.2  & 99.3 & -
    \\
    S-CNN+SVM * \cite{sharma2016hyperspectral}  & 99.4  & 99.6 & -
    \\
    Deep-Baseline  & {99.3} & {99.5} & {99.6}
    \\
    Deep-SSL  & \textbf{99.9} & \textbf{99.8}& \textbf{99.9}
    \\
     \hline
    \end{tabular}}
    \caption{Comparing accuracy (\%) of different band selection methods with our proposed method. }
    \label{tabb2}
\end{table}
\subsection{Comparison}
We compared our proposed algorithm with several existing face recognition techniques that are
extended to the hyperspectral face recognition methods. We categorize these methods into four groups including four existing hyperspectral face recognition methods \cite{robila2008toward,pan2009comparison,di2010studies,shen2012hyperspectral}, eight image-set classification methods \cite{kim2007discriminative,wang2008manifold,wang2009manifold,cevikalp2010face,hu2012face,wang2012covariance,uzair2015hyperspectral}
three RGB/grayscale face recognition algorithms \cite{wright2009robust,zhang2011sparse,lee2012local}, and one existing CNN-based model for hyperspectral  face  recognition \cite{sharma2016hyperspectral}.
For a fair comparison, we have been consistent with other compared methods in experimental setup including  the number of images in the gallery and probe data. Specifically, for the PolyU-HSFD dataset, we use the first 24 subjects which contain 113 hyperspectral image cubes. For each subject, we randomly select two cubes for the gallery and  we use the remaining 63 cubes for probes. For the CMU-HSFD dataset, the
dataset includes 48 subjects, each subject has 4 to 20 cubes
obtained from different sessions and different lighting
conditions. We use only the cubes which are obtained in a condition that all  lights are turned on. Thus, there are 147 hyperspectral cubes of 48 subjects such that each subject has 1 to 5 cubes. We construct the gallery  randomly by selecting one cube per subject, and we use  the remaining
99 cubes for probes. For the UWA-HSFD dataset, we randomly select one cube for each of 70 subjects to construct a gallery  
and we use the remaining 50 cubes for probes.

Table. \ref{tabb2} indicates the average accuracy of the compared methods when all  bands are available for different algorithms during the face recognition. The Deep-Baseline is the case where we use all the bands in our CNN framework for face recognition. Therefore, in this case we turn off the SSL  regularization term in (3), while Deep-SSL is the case that we perform face recognition using the SSL regularization term.  We reported the face recognition accuracy of Deep-SSL in Table. \ref{tabb2} to compare it with the best recognition results reported in the literature. The results show that Deep-SSL outperforms the state-of-the-art methods including PLS* and S-CNN+SVM* methods. The symbol * represents the case that the algorithms perform face recognition when they use their optimal hyperspectral bands. 

Please email us \footnote{ft0009@mix.wvu.edu or }  if you want to receive the data and the source code of our proposed algorithm presented in this book chapter.

\section{Conclusion}
In this work, we proposed a CNN-based model  which uses  an  SSL technique  to  select the  optimal  spectral  bands  to  obtain  the best  face  recognition  performance from all the spectral bands. In  this  method, convolutional filters in the first layer of our CNN are regularized by using a group Lasso algorithm to remove the redundant bands during the training. Experimental results indicate that our method automatically selects the optimal bands to obtain the best face recognition performance over that achieved using conventional broad-band (RGB) face images. Moreover, the results indicate that our model outperforms existing methods which also perform band selection for face recognition.
{\small \small
\bibliographystyle{ieeetr}
\bibliography{egbib}

\begin{thebibliography}{100}

\bibitem{allen2016overview}
D.~W. Allen, ``An overview of spectral imaging of human skin toward face
  recognition,'' in {\em Face Recognition Across the Imaging Spectrum},
  pp.~1--19, Springer, 2016.

\bibitem{pan2003face}
Z.~Pan, G.~Healey, M.~Prasad, and B.~Tromberg, ``Face recognition in
  hyperspectral images,'' {\em IEEE Transactions on Pattern Analysis and
  Machine Intelligence}, vol.~25, no.~12, pp.~1552--1560, 2003.

\bibitem{uzair2015hyperspectral}
M.~Uzair, A.~Mahmood, and A.~Mian, ``Hyperspectral face recognition with
  spatiospectral information fusion and {PLS} regression,'' {\em IEEE
  Transactions on Image Processing}, vol.~24, no.~3, pp.~1127--1137, 2015.

\bibitem{kruse2002comparison}
F.~A. Kruse {\em et~al.}, ``Comparison of {AVIRIS} and {H}yperion for
  hyperspectral mineral mapping,'' in {\em 11th JPL Airborne Geoscience
  Workshop}, vol.~4, 2002.

\bibitem{pan2009comparison}
Z.~Pan, G.~Healey, and B.~Tromberg, ``Comparison of spectral-only and
  spectral/spatial face recognition for personal identity verification,'' {\em
  EURASIP journal on Advances in Signal Processing}, vol.~2009, p.~8, 2009.

\bibitem{ryer2012quest}
D.~M. Ryer, T.~J. Bihl, K.~W. Bauer, and S.~K. Rogers, ``Quest hierarchy for
  hyperspectral face recognition,'' {\em Advances in Artificial Intelligence},
  vol.~2012, p.~1, 2012.

\bibitem{gross2001quo}
R.~Gross, J.~Shi, and J.~F. Cohn, {\em Quo vadis face recognition?}
\newblock Carnegie Mellon University, The Robotics Institute, 2001.

\bibitem{gross2004appearance}
R.~Gross, I.~Matthews, and S.~Baker, ``Appearance-based face recognition and
  light-fields,'' {\em IEEE Transactions on Pattern Analysis and Machine
  Intelligence}, vol.~26, no.~4, pp.~449--465, 2004.

\bibitem{martinez2002recognizing}
A.~M. Mart{\'\i}nez, ``Recognizing imprecisely localized, partially occluded,
  and expression variant faces from a single sample per class,'' {\em IEEE
  Transactions on Pattern Analysis \& Machine Intelligence}, no.~6,
  pp.~748--763, 2002.

\bibitem{wilder1996comparison}
J.~Wilder, P.~J. Phillips, C.~Jiang, and S.~Wiener, ``Comparison of visible and
  infra-red imagery for face recognition,'' in {\em Proceedings of the Second
  International Conference on Automatic Face and Gesture Recognition},
  pp.~182--187, IEEE, 1996.

\bibitem{blanz2002face}
V.~Blanz, S.~Romdhani, and T.~Vetter, ``Face identification across different
  poses and illuminations with a {3D} morphable model,'' in {\em Proceedings of
  Fifth IEEE International Conference on Automatic Face Gesture Recognition},
  pp.~202--207, IEEE, 2002.

\bibitem{anderson1981optics}
R.~R. Anderson and J.~A. Parrish, ``The optics of human skin,'' {\em Journal of
  investigative dermatology}, vol.~77, no.~1, pp.~13--19, 1981.

\bibitem{edwards1939pigments}
E.~A. Edwards and S.~Q. Duntley, ``The pigments and color of living human
  skin,'' {\em American Journal of Anatomy}, vol.~65, no.~1, pp.~1--33, 1939.

\bibitem{tsumura1999independent}
N.~Tsumura, H.~Haneishi, and Y.~Miyake, ``Independent-component analysis of
  skin color image,'' {\em JOSA A}, vol.~16, no.~9, pp.~2169--2176, 1999.

\bibitem{angelopoulo2001multispectral}
E.~Angelopoulo, R.~Molana, and K.~Daniilidis, ``Multispectral skin color
  modeling,'' in {\em Proceedings of the 2001 IEEE Computer Society Conference
  on Computer Vision and Pattern Recognition. CVPR 2001}, vol.~2, pp.~II--II,
  IEEE, 2001.

\bibitem{hagen2013review}
N.~A. Hagen and M.~W. Kudenov, ``Review of snapshot spectral imaging
  technologies,'' {\em Optical Engineering}, vol.~52, no.~9, p.~090901, 2013.

\bibitem{robila2008toward}
S.~A. Robila, ``Toward hyperspectral face recognition,'' in {\em Image
  Processing: Algorithms and Systems VI}, vol.~6812, p.~68120X, International
  Society for Optics and Photonics, 2008.

\bibitem{di2010studies}
W.~Di, L.~Zhang, D.~Zhang, and Q.~Pan, ``Studies on hyperspectral face
  recognition in visible spectrum with feature band selection,'' {\em IEEE
  Transactions on Systems, Man, and Cybernetics-Part A: Systems and Humans},
  vol.~40, no.~6, pp.~1354--1361, 2010.

\bibitem{shen2012hyperspectral}
L.~Shen and S.~Zheng, ``Hyperspectral face recognition using {3D} {G}abor
  wavelets,'' in {\em Pattern Recognition (ICPR), 2012 21st International
  Conference on}, pp.~1574--1577, IEEE, 2012.

\bibitem{bajcsy2004methodology}
P.~Bajcsy and P.~Groves, ``Methodology for hyperspectral band selection,'' {\em
  Photogrammetric Engineering \& Remote Sensing}, vol.~70, no.~7, pp.~793--802,
  2004.

\bibitem{chang1999joint}
C.-I. Chang, Q.~Du, T.-L. Sun, and M.~L. Althouse, ``A joint band
  prioritization and band-decorrelation approach to band selection for
  hyperspectral image classification,'' {\em IEEE transactions on geoscience
  and remote sensing}, vol.~37, no.~6, pp.~2631--2641, 1999.

\bibitem{melgani2004classification}
F.~Melgani and L.~Bruzzone, ``Classification of hyperspectral remote sensing
  images with support vector machines,'' {\em IEEE Transactions on geoscience
  and remote sensing}, vol.~42, no.~8, pp.~1778--1790, 2004.

\bibitem{keshava2001best}
N.~Keshava, ``Best bands selection for detection in hyperspectral processing,''
  in {\em Acoustics, Speech, and Signal Processing, 2001.
  Proceedings.(ICASSP'01). 2001 IEEE International Conference on}, vol.~5,
  pp.~3149--3152, IEEE, 2001.

\bibitem{du2003band}
Q.~Du, ``Band selection and its impact on target detection and classification
  in hyperspectral image analysis,'' in {\em Advances in Techniques for
  Analysis of Remotely Sensed Data, 2003 IEEE Workshop on}, pp.~374--377, IEEE,
  2003.

\bibitem{kaewpijit2003automatic}
S.~Kaewpijit, J.~Le~Moigne, and T.~El-Ghazawi, ``Automatic reduction of
  hyperspectral imagery using wavelet spectral analysis,'' {\em IEEE
  transactions on Geoscience and Remote Sensing}, vol.~41, no.~4, pp.~863--871,
  2003.

\bibitem{price1997spectral}
J.~C. Price, ``Spectral band selection for visible-near infrared remote
  sensing: spectral-spatial resolution tradeoffs,'' {\em IEEE Transactions on
  Geoscience and Remote Sensing}, vol.~35, no.~5, pp.~1277--1285, 1997.

\bibitem{steiner2016reliable}
H.~Steiner, A.~Kolb, and N.~Jung, ``Reliable face anti-spoofing using
  multispectral swir imaging,'' in {\em Biometrics (ICB), 2016 International
  Conference on}, pp.~1--8, IEEE, 2016.

\bibitem{bouchech2014dynamic}
H.~J. Bouchech, S.~Foufou, and M.~Abidi, ``Dynamic best spectral bands
  selection for face recognition,'' in {\em Information Sciences and Systems
  (CISS), 2014 48th Annual Conference on}, pp.~1--6, IEEE, 2014.

\bibitem{taherkhani2017restoring}
F.~Taherkhani and M.~Jamzad, ``Restoring highly corrupted images by impulse
  noise using radial basis functions interpolation,'' {\em IET Image
  Processing}, vol.~12, no.~1, pp.~20--30, 2017.

\bibitem{chen2013hyperspectral}
Y.~Chen, N.~M. Nasrabadi, and T.~D. Tran, ``Hyperspectral image classification
  via kernel sparse representation,'' {\em IEEE Transactions on Geoscience and
  Remote sensing}, vol.~51, no.~1, pp.~217--231, 2013.

\bibitem{talreja2019attribute}
V.~Talreja, F.~Taherkhani, M.~C. Valenti, and N.~M. Nasrabadi,
  ``Attribute-guided coupled gan for cross-resolution face recognition,'' {\em
  arXiv preprint arXiv:1908.01790}, 2019.

\bibitem{taherkhani2018facial}
F.~Taherkhani, V.~Talreja, H.~Kazemi, and N.~Nasrabadi, ``Facial attribute
  guided deep cross-modal hashing for face image retrieval,'' in {\em 2018
  International Conference of the Biometrics Special Interest Group (BIOSIG)},
  pp.~1--6, IEEE, 2018.

\bibitem{talreja2018using}
V.~Talreja, F.~Taherkhani, M.~C. Valenti, and N.~M. Nasrabadi, ``Using deep
  cross modal hashing and error correcting codes for improving the efficiency
  of attribute guided facial image retrieval,'' in {\em 2018 IEEE Global
  Conference on Signal and Information Processing (GlobalSIP)}, pp.~564--568,
  IEEE, 2018.

\bibitem{taherkhani2019matrix}
F.~Taherkhani, H.~Kazemi, and N.~M. Nasrabadi, ``Matrix completion for
  graph-based deep semi-supervised learning,'' in {\em Thirty-Third AAAI
  Conference on Artificial Intelligence}, 2019.

\bibitem{kazemi2018unsupervised}
H.~Kazemi, S.~Soleymani, F.~Taherkhani, S.~Iranmanesh, and N.~Nasrabadi,
  ``Unsupervised image-to-image translation using domain-specific variational
  information bound,'' in {\em Advances in Neural Information Processing
  Systems}, pp.~10369--10379, 2018.

\bibitem{kazemi2018unsupervised0}
H.~Kazemi, F.~Taherkhani, and N.~M. Nasrabadi, ``Unsupervised facial geometry
  learning for sketch to photo synthesis,'' in {\em 2018 International
  Conference of the Biometrics Special Interest Group (BIOSIG)}, pp.~1--5,
  IEEE, 2018.

\bibitem{talreja2017multibiometric}
V.~Talreja, M.~C. Valenti, and N.~M. Nasrabadi, ``Multibiometric secure system
  based on deep learning,'' in {\em 2017 IEEE Global conference on signal and
  information processing (globalSIP)}, pp.~298--302, IEEE, 2017.

\bibitem{talreja2018biometrics}
V.~Talreja, T.~Ferrett, M.~C. Valenti, and A.~Ross, ``Biometrics-as-a-service:
  A framework to promote innovative biometric recognition in the cloud,'' in
  {\em 2018 IEEE International Conference on Consumer Electronics (ICCE)},
  pp.~1--6, IEEE, 2018.

\bibitem{he_resnet_2016}
K.~He, X.~Zhang, S.~Ren, and J.~Sun, ``Deep residual learning for image
  recognition,'' in {\em Proc. IEEE Conference on Computer Vision and Pattern
  Recognition}, pp.~770--778, June 2016.

\bibitem{krizhevsky_imagenet_2012}
A.~Krizhevsky, I.~Sutskever, and G.~E. Hinton, ``Imagenet classification with
  deep convolutional neural networks,'' in {\em Proc. Advances in Neural
  Information Processing Systems}, pp.~1097--1105, Dec. 2012.

\bibitem{simonyan_very_deep_2014}
K.~Simonyan and A.~Zisserman, ``Very deep convolutional networks for
  large-scale image recognition,'' {\em CoRR}, vol.~abs/1409.1556, Sept. 2014.

\bibitem{Erhan_2014_CVPR}
D.~Erhan, C.~Szegedy, A.~Toshev, and D.~Anguelov, ``Scalable object detection
  using deep neural networks,'' in {\em Proc. IEEE Conference on Computer
  Vision and Pattern Recognition}, June 2014.

\bibitem{ren_faster_rcnn_2015}
S.~Ren, K.~He, R.~Girshick, and J.~Sun, ``Faster r-cnn: Towards real-time
  object detection with region proposal networks,'' in {\em Proc. Advances in
  Neural Information Processing Systems}, pp.~91--99, Dec. 2015.

\bibitem{soleymani2019defending}
S.~Soleymani, A.~Dabouei, J.~Dawson, and N.~M. Nasrabadi, ``Defending against
  adversarial iris examples using wavelet decomposition,'' {\em arXiv preprint
  arXiv:1908.03176}, 2019.

\bibitem{soleymani2018prosodic}
S.~Soleymani, A.~Dabouei, S.~M. Iranmanesh, H.~Kazemi, J.~Dawson, and N.~M.
  Nasrabadi, ``Prosodic-enhanced siamese convolutional neural networks for
  cross-device text-independent speaker verification,'' in {\em 2018 IEEE 9th
  International Conference on Biometrics Theory, Applications and Systems
  (BTAS)}, pp.~1--7, IEEE, 2018.

\bibitem{soleymani2019adversarial}
S.~Soleymani, A.~Dabouei, J.~Dawson, and N.~M. Nasrabadi, ``Adversarial
  examples to fool iris recognition systems,'' {\em arXiv preprint
  arXiv:1906.09300}, 2019.

\bibitem{silver_2016_mastering}
D.~Silver, A.~Huang, C.~J. Maddison, A.~Guez, L.~Sifre, G.~Van Den~Driessche,
  J.~Schrittwieser, I.~Antonoglou, V.~Panneershelvam, M.~Lanctot, {\em et~al.},
  ``Mastering the game of go with deep neural networks and tree search,'' {\em
  Nature}, vol.~529, p.~484, Jan. 2016.

\bibitem{zhong2017deep}
Z.~Zhong, J.~Li, L.~Ma, H.~Jiang, and H.~Zhao, ``Deep residual networks for
  hyperspectral image classification,'' in {\em Geoscience and Remote Sensing
  Symposium (IGARSS), 2017 IEEE International}, pp.~1824--1827, IEEE, 2017.

\bibitem{zhan2017hyperspectral}
Y.~Zhan, D.~Hu, H.~Xing, and X.~Yu, ``Hyperspectral band selection based on
  deep convolutional neural network and distance density,'' {\em IEEE
  Geoscience and Remote Sensing Letters}, vol.~14, no.~12, pp.~2365--2369,
  2017.

\bibitem{zhan2017new}
Y.~Zhan, H.~Tian, W.~Liu, Z.~Yang, K.~Wu, G.~Wang, P.~Chen, and X.~Yu, ``A new
  hyperspectral band selection approach based on convolutional neural
  network,'' in {\em Geoscience and Remote Sensing Symposium (IGARSS), 2017
  IEEE International}, pp.~3660--3663, IEEE, 2017.

\bibitem{sharma2016hyperspectral}
V.~Sharma, A.~Diba, T.~Tuytelaars, and L.~Van~Gool, ``Hyperspectral cnn for
  image classification \& band selection, with application to face
  recognition,'' 2016.

\bibitem{li2018novel}
N.~Li, C.~Wang, H.~Zhao, X.~Gong, and D.~Wang, ``A novel deep convolutional
  neural network for spectral-spatial classification of hyperspectral data.,''
  {\em International Archives of the Photogrammetry, Remote Sensing \& Spatial
  Information Sciences}, vol.~42, no.~3, 2018.

\bibitem{guo2012feature}
Z.~Guo, D.~Zhang, L.~Zhang, and W.~Liu, ``Feature band selection for online
  multispectral palmprint recognition,'' {\em IEEE Transactions on Information
  Forensics and Security}, vol.~7, no.~3, pp.~1094--1099, 2012.

\bibitem{yuan2006model}
M.~Yuan and Y.~Lin, ``Model selection and estimation in regression with grouped
  variables,'' {\em Journal of the Royal Statistical Society: Series B
  (Statistical Methodology)}, vol.~68, no.~1, pp.~49--67, 2006.

\bibitem{hanson1989comparing}
S.~J. Hanson and L.~Y. Pratt, ``Comparing biases for minimal network
  construction with back-propagation,'' in {\em Advances in neural information
  processing systems}, pp.~177--185, 1989.

\bibitem{hassibi1993second}
B.~Hassibi and D.~G. Stork, ``Second order derivatives for network pruning:
  Optimal brain surgeon,'' in {\em Advances in neural information processing
  systems}, pp.~164--171, 1993.

\bibitem{han2015learning}
S.~Han, J.~Pool, J.~Tran, and W.~Dally, ``Learning both weights and connections
  for efficient neural network,'' in {\em Advances in neural information
  processing systems}, pp.~1135--1143, 2015.

\bibitem{wen2016learning}
W.~Wen, C.~Wu, Y.~Wang, Y.~Chen, and H.~Li, ``Learning structured sparsity in
  deep neural networks,'' in {\em Advances in neural information processing
  systems}, pp.~2074--2082, 2016.

\bibitem{li2016pruning}
H.~Li, A.~Kadav, I.~Durdanovic, H.~Samet, and H.~P. Graf, ``Pruning filters for
  efficient convnets,'' {\em arXiv preprint arXiv:1608.08710}, 2016.

\bibitem{murray2015auto}
K.~Murray and D.~Chiang, ``Auto-sizing neural networks: With applications to
  n-gram language models,'' {\em arXiv preprint arXiv:1508.05051}, 2015.

\bibitem{feng2015learning}
J.~Feng and T.~Darrell, ``Learning the structure of deep convolutional
  networks,'' in {\em Proceedings of the IEEE international conference on
  computer vision}, pp.~2749--2757, 2015.

\bibitem{griffiths2011indian}
T.~L. Griffiths and Z.~Ghahramani, ``The indian buffet process: An introduction
  and review,'' {\em Journal of Machine Learning Research}, vol.~12, no.~Apr,
  pp.~1185--1224, 2011.

\bibitem{hu2016network}
H.~Hu, R.~Peng, Y.-W. Tai, and C.-K. Tang, ``Network trimming: A data-driven
  neuron pruning approach towards efficient deep architectures,'' {\em arXiv
  preprint arXiv:1607.03250}, 2016.

\bibitem{anwar2017structured}
S.~Anwar, K.~Hwang, and W.~Sung, ``Structured pruning of deep convolutional
  neural networks,'' {\em ACM Journal on Emerging Technologies in Computing
  Systems (JETC)}, vol.~13, no.~3, p.~32, 2017.

\bibitem{lecun1990optimal}
Y.~LeCun, J.~S. Denker, and S.~A. Solla, ``Optimal brain damage,'' in {\em
  Advances in neural information processing systems}, pp.~598--605, 1990.

\bibitem{cirecsan2011high}
D.~C. Cire{\c{s}}an, U.~Meier, J.~Masci, L.~M. Gambardella, and J.~Schmidhuber,
  ``High-performance neural networks for visual object classification,'' {\em
  arXiv preprint arXiv:1102.0183}, 2011.

\bibitem{ioannou2017deep}
Y.~Ioannou, D.~Robertson, R.~Cipolla, and A.~Criminisi, ``Deep roots: Improving
  cnn efficiency with hierarchical filter groups,'' in {\em Proceedings of the
  IEEE Conference on Computer Vision and Pattern Recognition}, pp.~1231--1240,
  2017.

\bibitem{howard2017mobilenets}
A.~G. Howard, M.~Zhu, B.~Chen, D.~Kalenichenko, W.~Wang, T.~Weyand,
  M.~Andreetto, and H.~Adam, ``Mobilenets: Efficient convolutional neural
  networks for mobile vision applications,'' {\em arXiv preprint
  arXiv:1704.04861}, 2017.

\bibitem{arora2014provable}
S.~Arora, A.~Bhaskara, R.~Ge, and T.~Ma, ``Provable bounds for learning some
  deep representations,'' in {\em International Conference on Machine
  Learning}, pp.~584--592, 2014.

\bibitem{gong2014compressing}
Y.~Gong, L.~Liu, M.~Yang, and L.~Bourdev, ``Compressing deep convolutional
  networks using vector quantization,'' {\em arXiv preprint arXiv:1412.6115},
  2014.

\bibitem{anwar2015fixed}
S.~Anwar, K.~Hwang, and W.~Sung, ``Fixed point optimization of deep
  convolutional neural networks for object recognition,'' in {\em 2015 IEEE
  International Conference on Acoustics, Speech and Signal Processing
  (ICASSP)}, pp.~1131--1135, IEEE, 2015.

\bibitem{chen2015compressing}
W.~Chen, J.~Wilson, S.~Tyree, K.~Weinberger, and Y.~Chen, ``Compressing neural
  networks with the hashing trick,'' in {\em International Conference on
  Machine Learning}, pp.~2285--2294, 2015.

\bibitem{han2016dsd}
S.~Han, J.~Pool, S.~Narang, H.~Mao, S.~Tang, E.~Elsen, B.~Catanzaro, J.~Tran,
  and W.~J. Dally, ``{DSD}: regularizing deep neural networks with
  dense-sparse-dense training flow,'' {\em arXiv preprint arXiv:1607.04381},
  vol.~3, no.~6, 2016.

\bibitem{denton2014exploiting}
E.~L. Denton, W.~Zaremba, J.~Bruna, Y.~LeCun, and R.~Fergus, ``Exploiting
  linear structure within convolutional networks for efficient evaluation,'' in
  {\em Advances in neural information processing systems}, pp.~1269--1277,
  2014.

\bibitem{jaderberg2014speeding}
M.~Jaderberg, A.~Vedaldi, and A.~Zisserman, ``Speeding up convolutional neural
  networks with low rank expansions,'' {\em arXiv preprint arXiv:1405.3866},
  2014.

\bibitem{lebedev2014speeding}
V.~Lebedev, Y.~Ganin, M.~Rakhuba, I.~Oseledets, and V.~Lempitsky, ``Speeding-up
  convolutional neural networks using fine-tuned {CP}-decomposition,'' {\em
  arXiv preprint arXiv:1412.6553}, 2014.

\bibitem{zhang2016accelerating}
X.~Zhang, J.~Zou, K.~He, and J.~Sun, ``Accelerating very deep convolutional
  networks for classification and detection,'' {\em IEEE transactions on
  pattern analysis and machine intelligence}, vol.~38, no.~10, pp.~1943--1955,
  2016.

\bibitem{kim2015compression}
Y.-D. Kim, E.~Park, S.~Yoo, T.~Choi, L.~Yang, and D.~Shin, ``Compression of
  deep convolutional neural networks for fast and low power mobile
  applications,'' {\em arXiv preprint arXiv:1511.06530}, 2015.

\bibitem{wan2013regularization}
L.~Wan, M.~Zeiler, S.~Zhang, Y.~Le~Cun, and R.~Fergus, ``Regularization of
  neural networks using dropconnect,'' in {\em International conference on
  machine learning}, pp.~1058--1066.

\bibitem{jin2016training}
X.~Jin, X.~Yuan, J.~Feng, and S.~Yan, ``Training skinny deep neural networks
  with iterative hard thresholding methods,'' {\em arXiv preprint
  arXiv:1607.05423}, 2016.

\bibitem{fan2010human}
J.~Fan, W.~Xu, Y.~Wu, and Y.~Gong, ``Human tracking using convolutional neural
  networks,'' {\em IEEE Transactions on Neural Networks}, vol.~21, no.~10,
  pp.~1610--1623, 2010.

\bibitem{toshev2014deeppose}
A.~Toshev and C.~Szegedy, ``Deeppose: Human pose estimation via deep neural
  networks,'' in {\em Proceedings of the IEEE conference on computer vision and
  pattern recognition}, pp.~1653--1660, 2014.

\bibitem{zhao2015saliency}
R.~Zhao, W.~Ouyang, H.~Li, and X.~Wang, ``Saliency detection by multi-context
  deep learning,'' in {\em Proceedings of the IEEE Conference on Computer
  Vision and Pattern Recognition}, pp.~1265--1274, 2015.

\bibitem{donahue2014decaf}
J.~Donahue, Y.~Jia, O.~Vinyals, J.~Hoffman, N.~Zhang, E.~Tzeng, and T.~Darrell,
  ``Decaf: A deep convolutional activation feature for generic visual
  recognition,'' in {\em International conference on machine learning},
  pp.~647--655, 2014.

\bibitem{zhao2019object}
Z.-Q. Zhao, P.~Zheng, S.-t. Xu, and X.~Wu, ``Object detection with deep
  learning: A review,'' {\em IEEE transactions on neural networks and learning
  systems}, 2019.

\bibitem{simonyan2014very}
K.~Simonyan and A.~Zisserman, ``Very deep convolutional networks for
  large-scale image recognition,'' {\em arXiv preprint arXiv:1409.1556}, 2014.

\bibitem{zeiler2014visualizing}
M.~D. Zeiler and R.~Fergus, ``Visualizing and understanding convolutional
  networks,'' in {\em European conference on computer vision}, pp.~818--833,
  Springer, 2014.

\bibitem{zeiler2013stochastic}
M.~D. Zeiler and R.~Fergus, ``Stochastic pooling for regularization of deep
  convolutional neural networks,'' {\em arXiv preprint arXiv:1301.3557}, 2013.

\bibitem{rippel2015spectral}
O.~Rippel, J.~Snoek, and R.~P. Adams, ``Spectral representations for
  convolutional neural networks,'' in {\em Advances in neural information
  processing systems}, pp.~2449--2457, 2015.

\bibitem{nguyen2015deep}
A.~Nguyen, J.~Yosinski, and J.~Clune, ``Deep neural networks are easily fooled:
  High confidence predictions for unrecognizable images,'' in {\em Proceedings
  of the IEEE conference on computer vision and pattern recognition},
  pp.~427--436, 2015.

\bibitem{gong2014multi}
Y.~Gong, L.~Wang, R.~Guo, and S.~Lazebnik, ``Multi-scale orderless pooling of
  deep convolutional activation features,'' in {\em European conference on
  computer vision}, pp.~392--407, Springer, 2014.

\bibitem{springenberg2014striving}
J.~T. Springenberg, A.~Dosovitskiy, T.~Brox, and M.~Riedmiller, ``Striving for
  simplicity: The all convolutional net,'' {\em arXiv preprint
  arXiv:1412.6806}, 2014.

\bibitem{zagoruyko2016wide}
S.~Zagoruyko and N.~Komodakis, ``Wide residual networks,'' {\em arXiv preprint
  arXiv:1605.07146}, 2016.

\bibitem{krizhevsky2012imagenet}
A.~Krizhevsky, I.~Sutskever, and G.~E. Hinton, ``Imagenet classification with
  deep convolutional neural networks,'' in {\em Advances in neural information
  processing systems}, pp.~1097--1105, 2012.

\bibitem{wen2016discriminative}
Y.~Wen, K.~Zhang, Z.~Li, and Y.~Qiao, ``A discriminative feature learning
  approach for deep face recognition,'' in {\em European Conference on Computer
  Vision}, pp.~499--515, Springer, 2016.

\bibitem{yi2014learning}
D.~Yi, Z.~Lei, S.~Liao, and S.~Z. Li, ``Learning face representation from
  scratch,'' {\em arXiv preprint arXiv:1411.7923}, 2014.

\bibitem{kingma2014adam}
D.~P. Kingma and J.~Ba, ``Adam: A method for stochastic optimization,'' {\em
  arXiv preprint arXiv:1412.6980}, 2014.

\bibitem{denes2002hyperspectral}
L.~J. Denes, P.~Metes, and Y.~Liu, {\em Hyperspectral face database}.
\newblock Carnegie Mellon University, The Robotics Institute, 2002.

\bibitem{uzair2013hyperspectral}
M.~Uzair, A.~Mahmood, and A.~S. Mian, ``Hyperspectral face recognition using
  {3D-DCT} and partial least squares.,'' in {\em BMVC}, 2013.

\bibitem{kim2007discriminative}
T.-K. Kim, J.~Kittler, and R.~Cipolla, ``Discriminative learning and
  recognition of image set classes using canonical correlations,'' {\em IEEE
  Transactions on Pattern Analysis and Machine Intelligence}, vol.~29, no.~6,
  pp.~1005--1018, 2007.

\bibitem{wang2008manifold}
R.~Wang, S.~Shan, X.~Chen, and W.~Gao, ``Manifold-manifold distance with
  application to face recognition based on image set,'' in {\em Computer Vision
  and Pattern Recognition, 2008. CVPR 2008. IEEE Conference on}, pp.~1--8,
  IEEE, 2008.

\bibitem{wang2009manifold}
R.~Wang and X.~Chen, ``Manifold discriminant analysis,'' in {\em Computer
  Vision and Pattern Recognition, 2009. CVPR 2009. IEEE Conference on},
  pp.~429--436, IEEE, 2009.

\bibitem{cevikalp2010face}
H.~Cevikalp and B.~Triggs, ``Face recognition based on image sets,'' in {\em
  Computer Vision and Pattern Recognition (CVPR), 2010 IEEE Conference on},
  pp.~2567--2573, IEEE, 2010.

\bibitem{hu2012face}
Y.~Hu, A.~S. Mian, and R.~Owens, ``Face recognition using sparse approximated
  nearest points between image sets,'' {\em IEEE transactions on pattern
  analysis and machine intelligence}, vol.~34, no.~10, pp.~1992--2004, 2012.

\bibitem{wang2012covariance}
R.~Wang, H.~Guo, L.~S. Davis, and Q.~Dai, ``Covariance discriminative learning:
  A natural and efficient approach to image set classification,'' in {\em
  Computer Vision and Pattern Recognition (CVPR), 2012 IEEE Conference on},
  pp.~2496--2503, IEEE, 2012.

\bibitem{wright2009robust}
J.~Wright, A.~Y. Yang, A.~Ganesh, S.~S. Sastry, and Y.~Ma, ``Robust face
  recognition via sparse representation,'' {\em IEEE transactions on pattern
  analysis and machine intelligence}, vol.~31, no.~2, pp.~210--227, 2009.

\bibitem{zhang2011sparse}
L.~Zhang, M.~Yang, and X.~Feng, ``Sparse representation or collaborative
  representation: Which helps face recognition?,'' in {\em Computer vision
  (ICCV), 2011 IEEE international conference on}, pp.~471--478, IEEE, 2011.

\bibitem{lee2012local}
S.~H. Lee, J.~Y. Choi, Y.~M. Ro, and K.~N. Plataniotis, ``Local color vector
  binary patterns from multichannel face images for face recognition,'' {\em
  IEEE Transactions on Image Processing}, vol.~21, no.~4, pp.~2347--2353, 2012.

\end{thebibliography}
}

\end{document}